\documentclass[10pt,twocolumn,letterpaper]{article}

\usepackage{iccv}                %

\usepackage{multirow}
\usepackage{multicol}
\usepackage{colortbl}
\usepackage{subcaption}
\usepackage{subfloat}

\usepackage[accsupp]{axessibility}  %

\newcommand{\frameworkName}{\mbox{CoMPaSS}}

\newcommand{\dataEngineName}{\mbox{SCOP}}

\newcommand{\peName}{\mbox{TENOR}}

\newcommand{\q}[1]{``#1''}
\newcommand{\vv}[1]{\q{\texttt{#1}}}

\newcommand{\isbest}[1]{\textbf{#1}}

\newcommand{\ispriorbest}[1]{#1}

\definecolor{iccvblue}{rgb}{0.21,0.49,0.74}
\usepackage[pagebackref,breaklinks,colorlinks,allcolors=iccvblue]{hyperref}

\def\paperID{6837} %
\def\confName{ICCV}
\def\confYear{2025}

\title{
  \frameworkName{}:
  Enhancing Spatial Understanding in Text-to-Image Diffusion Models
}

\author{%
  \begin{tabular}{@{}>{\centering\arraybackslash}p{0.22\textwidth}@{\hspace{0.02\textwidth}}>{\centering\arraybackslash}p{0.22\textwidth}@{\hspace{0.02\textwidth}}>{\centering\arraybackslash}p{0.22\textwidth}@{\hspace{0.02\textwidth}}>{\centering\arraybackslash}p{0.22\textwidth}@{}}
    Gaoyang Zhang\textsuperscript{1}\thanks{Work done during internship at vivo.} & Bingtao Fu\textsuperscript{3} & Qingnan Fan\textsuperscript{2} & Qi Zhang\textsuperscript{2} \\
  {\tt\small blurgy@zju.edu.cn} & {\tt\small bingtaofu93@gmail.com} & {\tt\small fqnchina@gmail.com} & {\tt\small nwpuqzhang@gmail.com} \\[1em]
  Runxing Liu\textsuperscript{2} & Hong Gu\textsuperscript{2} & Huaqi Zhang\textsuperscript{2} & Xinguo Liu\textsuperscript{1}\thanks{Corresponding author.} \\
  {\tt\small runxingliu13@gmail.com} & {\tt\small guhong@vivo.com} & {\tt\small zhanghuaqi@vivo.com} & {\tt\small xinguoliu@zju.edu.cn}
  \end{tabular} \\[3em]
  \textsuperscript{1}Zhejiang University, China
  \qquad
  \textsuperscript{2}vivo, China
  \qquad
  \textsuperscript{3}Ant Group, China
}

\begin{document}
\maketitle
\begin{abstract}
  Text-to-image (T2I) diffusion models excel at generating photorealistic
  images but often fail to render accurate spatial relationships.
  We identify two core issues underlying this common failure:
  1) the ambiguous nature of data concerning spatial relationships in existing
     datasets, and
  2) the inability of current text encoders to accurately interpret the
     spatial semantics of input descriptions.
  We propose \frameworkName{}, a versatile framework that enhances spatial
  understanding in T2I models.
  It first addresses data ambiguity with the \textbf{S}patial
  \textbf{C}onstraints-\textbf{O}riented \textbf{P}airing (\dataEngineName{})
  data engine, which curates spatially-accurate training data via principled
  constraints.
  To leverage these priors, \frameworkName{} also introduces the
  \textbf{T}oken \textbf{EN}coding \textbf{OR}dering (\peName{}) module, which
  preserves crucial token ordering information lost by text encoders, thereby
  reinforcing the prompt's linguistic structure.
  Extensive experiments on four popular T2I models (UNet and MMDiT-based) show
  \frameworkName{} sets a new state of the art on key spatial benchmarks, with
  substantial relative gains on VISOR (+98\%), T2I-CompBench Spatial (+67\%),
  and GenEval Position (+131\%).
  Code is available at \href{https://github.com/blurgyy/CoMPaSS}
  {\texttt{github.com/blurgyy/CoMPaSS}}.
\end{abstract}

\section{Introduction}

Recent advancements in text-to-image (T2I) diffusion
models~\cite{saharia_photorealistic_2022, ramesh_hierarchical_2022,
rombach_high-resolution_2022} have transformed visual content creation and
significantly shaped the modern digital life~\cite{betker_improving_nodate,
MidJourney, RunwayML, BlackForestLabs2024FLUX1dev}.
These models are capable of synthesizing photorealistic images with remarkable
details~\cite{dhariwal_diffusion_2021}, and are now widely applied across a
range of creative and practical tasks~\cite{hertz_prompt--prompt_2022,
lugmayr_repaint_2022, ruiz_dreambooth_2023, poole_dreamfusion_2022}.
Despite these advancements, a key limitation persists: today's diffusion
models frequently fail to correctly render spatial relationships described in
text.
For example, when given seemingly simple spatial configurations like \vv{a
motorcycle to the right of a bear}, or \vv{a bird below a skateboard}, models
that excel in realism often struggle to accurately capture these spatial
relationships.
Given that T2I diffusion models have shown significant capability in encoding
abstract attributes such as aesthetic style and
quality~\cite{dhariwal_diffusion_2021, rombach_high-resolution_2022,
podell_sdxl_2023, esser_scaling_2024, BlackForestLabs2024FLUX1dev},
their consistent failure in handling spatial relationships warrants closer
examination.

\begin{figure}
  \centering
  \includegraphics[width=.985\linewidth]{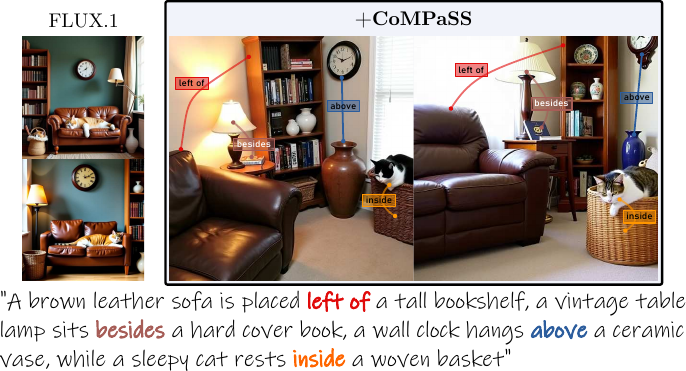}
  \caption{
    \textbf{\frameworkName{}} enhances the spatial understanding of existing T2I
    diffusion models, enabling them to generate images that faithfully reflect
    spatial configurations specified in the text prompt.
  } \label{fig@teaser}
\end{figure}

To better understand this limitation, we conduct a thorough investigation
covering the essential aspects of training a diffusion model, including the
characteristics of the training data and the quality of language
representations used during generation.
Our analysis identified two core issues underlying this spatial understanding
deficiency.
\textbf{First}, the organization of data concerning spatial relationships in
widely-used image-text datasets including
LAION~\cite{schuhmann_laion-400m_2021, schuhmann_laion-5b_2022},
CC12M~\cite{changpinyo_conceptual_2021}, and COCO~\cite{lin_microsoft_2014}
is severely flawed.
As shown in \cref{fig@dataset-flaw}, spatial descriptions in these datasets
are frequently ambiguous due to various reasons, such as inconsistent frames
of reference or the non-spatial use of spatial terms.
This inevitably leads to the spatial terms being interpreted by the diffusion
model as ambiguous.
\textbf{Second}, commonly used text encoders including
CLIP~\cite{dblp-clip, dblp-cherti_reproducible_2023} and
T5~\cite{raffel_exploring_2020} consistently fail to preserve spatial
relationships in their embeddings.
We demonstrate this through a proxy task (\cref{tab@text-encoder-flaw}) where
we test the ability of text encoders to identify spatially equivalent
descriptions among a set of prompt variations. Our results show that these
encoders struggle to interpret spatial descriptions accurately.
Furthermore, the two issues compound each other.
Even if a text encoder could reliably capture spatial relationships, the lack
of consistent and accurate spatial data would still limit model performance.
Conversely, even high-quality spatial descriptions would be undermined by the
encoder’s inability to interpret them accurately.

To this end, we propose \frameworkName{} (Comprehensive Method for Positional
and Spatial Synthesis), a comprehensive solution to enhance the spatial
understanding of T2I diffusion models.
Central to our approach is the Spatial Constraints-Oriented Pairing
(\dataEngineName{}) data engine, which extracts object pairs with clear
spatial relationships from images, along with their accurate textual
descriptions.
By enforcing strict criteria for visual prominence, positional clarity, and
size balance, \dataEngineName{} identifies and validates spatial relationships
between object pairs through carefully designed spatial constraints.
When applied to the COCO~\cite{lin_microsoft_2014} training split,
\dataEngineName{} yields a curated set of over 28,000 object pairs, each
coupled with spatially-accurate text descriptions.
To better exploit the spatial priors curated by \dataEngineName{}, we
introduce the \mbox{Token} \mbox{ENcoding} \mbox{ORdering} (\peName{}) module.
This plug-and-play module addresses a key limitation of text encoders by
reinforcing the prompt's linguistic structure.
It does so by injecting explicit token ordering information into the model's
attention mechanism, ensuring that the sequential nature of the text is
preserved.
\peName{} is compatible with any diffusion architecture, adds no extra
parameters, and incurs negligible inference-time overhead.

When applied to popular open-weight T2I diffusion models (SD1.4, SD1.5, SD2.1,
and FLUX.1), \frameworkName{} achieves state-of-the-art performance on
well-known benchmarks, with substantial relative gains on
VISOR~\cite{gokhale_benchmarking_2023} (+98\%), T2I-CompBench
Spatial~\cite{dblp-t2i-compbench} (+67\%), and GenEval
Position~\cite{dblp-geneval} (+131\%).
Notably, these improvements are achieved without compromising general
generation capabilities, while also enhancing image fidelity.
In summary, our contributions are three-fold:

\begin{itemize}
  \item
    A comprehensive training framework, \frameworkName{}, that significantly
    enhances spatial understanding in T2I diffusion models.

  \item
    A systematic data engine, \dataEngineName{}, that enforces principled
    constraints to identify and validate unambiguous spatial relationships
    between object pairs in images, enabling the curation of high-quality
    spatial training data.

  \item
    A parameter-free module, \peName{}, that improves a model's ability to
    interpret spatial language by preserving the crucial token order of text
    prompts, while adding negligible computational overhead.
\end{itemize}

\section{Related Work}

\textbf{Text-to-image Diffusion Models.}
Diffusion models have achieved very high-quality and diverse image synthesis
in recent years~\cite{ho_denoising_2020, rombach_high-resolution_2022,
dhariwal_diffusion_2021, peebles_scalable_2023, dblp-pixart-alpha,
chen_pixart-_2024, dblp-pixart-sigma}.
To enable user input via natural language, text-conditional diffusion models
are trained on large-scale image-text
datasets~\cite{changpinyo_conceptual_2021, schuhmann_laion-400m_2021,
schuhmann_laion-5b_2022, liang_rich_2024}, typically using pre-trained text
encoders to bridge the two modalities.
While early models such as GLIDE~\cite{nichol_glide_2022},
\mbox{DALLE-2}~\cite{ramesh_hierarchical_2022}, and the Stable Diffusion
family~\cite{rombach_high-resolution_2022, podell_sdxl_2023,
esser_scaling_2024} employed CLIP~\cite{dblp-clip,
dblp-cherti_reproducible_2023} for language
conditioning, more recent work like PixArt~\cite{dblp-pixart-alpha,
chen_pixart-_2024, dblp-pixart-sigma}, SD3~\cite{esser_scaling_2024}, and
FLUX.1 has adopted text-only encoders like T5~\cite{raffel_exploring_2020} to
further enhance faithfulness to the input prompt.
To quantify this faithfulness, benchmarks with diverse evaluation
methodologies have been explored~\cite{hu_tifa_2023, dblp-geneval,
gokhale_benchmarking_2023, dblp-t2i-compbench, dblp-dpg-bench,
li_genai-bench_2024, feng_tc-bench_2024, wang_diffusiondb_2023,
saharia_photorealistic_2022, yu_scaling_2022}.
Although recent reinforcement learning-based methods~\cite{fan_dpok_2023,
saeidi_dual_2025, karthik_scalable_2024, wallace_diffusion_2023,
black_training_2024, gu_diffusion-rpo_2024, dblp-t2i-compbench}, which often
optimize for human preference using specialized datasets~\cite{liang_rich_2024,
segalis_picture_2023, bakr_hrs-bench_2023, wu_human_2023-2, wu_human_2023-1,
xu_imagereward_2023, dblp-t2i-compbench}, have exhibited strong benchmark
scores, the ambiguity underlying these datasets remains a significant
challenge.
Our work addresses this issue by introducing a data engine designed to ensure
accurate spatial alignment between images and text in the training data.

\noindent\textbf{Spatial Control in Diffusion Models.}
Spatial control is one of the most popular demands in image
generation~\cite{gucluturk_convolutional_2016, sangkloy_scribbler_2017,
zhao_image_2019, harkonen_ganspace_2020, wu_stylespace_2021, pan_drag_2023}.
Approaches for controllable text-to-image generation can be broadly
categorized by whether they manipulate the training or the inference stage of
a diffusion model.
Training-based methods~\cite{ruiz_dreambooth_2023, zhang_adding_2023,
mou_t2i-adapter_2023, ye_ip-adapter_2023, dblp-spright, wang_tokencompose_2024}
adapt the pretrained weights of diffusion models for specific downstream
tasks~\cite{xiao_fastcomposer_2023, li_photomaker_2023, wang_instantid_2024,
yu_scaling_2024, wu_seesr_2023, wang_tokencompose_2024, dblp-spright,
dblp-dpg-bench}.
However, they generally incur significant training overhead due to
architectural augmentations~\cite{zhang_adding_2023, dblp-dpg-bench,
wu_paragraph--image_2023}, extra training modules~\cite{zhang_adding_2023,
dblp-spright}, or the need for additional
supervision~\cite{wang_tokencompose_2024, jiang_comat_2024}.
Inference-only methods~\cite{dblp-layout-guidance, dblp-rnb,
dblp-attention-refocusing, liu_compositional_2023, feng_training-free_2023,
chefer_attend-and-excite_2023, hertz_prompt--prompt_2022, taghipour_box_2024,
lv_pick-and-draw_2024} achieve robust targeted generation and
editing, but often rely on explicit inputs like bounding boxes. These inputs
may require manual annotation or specialist models that have separate training
costs~\cite{feng_layoutgpt_2024, cho_visual_2023} or high inference
costs~\cite{lian_llm-grounded_2023, dblp-rnb}.
Furthermore, many of these approaches introduce significant inference overhead
due to costly test-time optimization.
In contrast, our work introduces a two-part solution.
Our \dataEngineName{} data engine provides high-quality training data, while
our \peName{} module ensures the model can leverage this data by preserving
the crucial token order of text prompts.
This combined approach efficiently improves spatial understanding without
extra trainable parameters or significant computational overhead.

\section{Approach}

We present \frameworkName{}, a comprehensive solution for improving spatial
understanding in text-to-image diffusion models.
In \cref{sec@scope}, we introduce the \mbox{Spatial}
\mbox{Constraints}-\mbox{Oriented} \mbox{Pairing} (\dataEngineName{}) data
engine, which extracts object pairs with clear spatial relationships from
images, along with their accurate textual descriptions.
In \cref{sec@tenor}, we introduce the \mbox{Token} \mbox{ENcoding}
\mbox{ORdering} (\peName{}) module, a plug-and-play module that improves the
model's ability to process spatial language by preserving the crucial token
order of text prompts.

These two components work in synergy:
\dataEngineName{} provides high-quality spatial priors through carefully
curated spatial relationships, while
\peName{} ensures the model can distinguish between structurally different
prompts, allowing it to better leverage these spatial priors during
generation.

\subsection{The \dataEngineName{} Data Engine} \label{sec@scope}

High-quality spatial understanding in generative models requires training data
with unambiguous spatial relationships.
However, existing text-image datasets often contain problematic spatial
descriptions.
As shown in \cref{fig@dataset-flaw},
1) terms like \vv{left} and \vv{right} suffer from perspective ambiguity,
   referring inconsistently to viewer perspective or object-intrinsic
   orientation;
2) directional terms are frequently used in non-spatial contexts (\eg{}
   \vv{the right choice});
3) spatial relationships are often described with missing or incorrect
   reference objects, making the intended spatial configuration impossible to
   determine.
These issues create significant challenges for models attempting to learn
reliable spatial relationships from such data.

\begin{figure*}
  \centering
  \includegraphics[width=.93\textwidth]{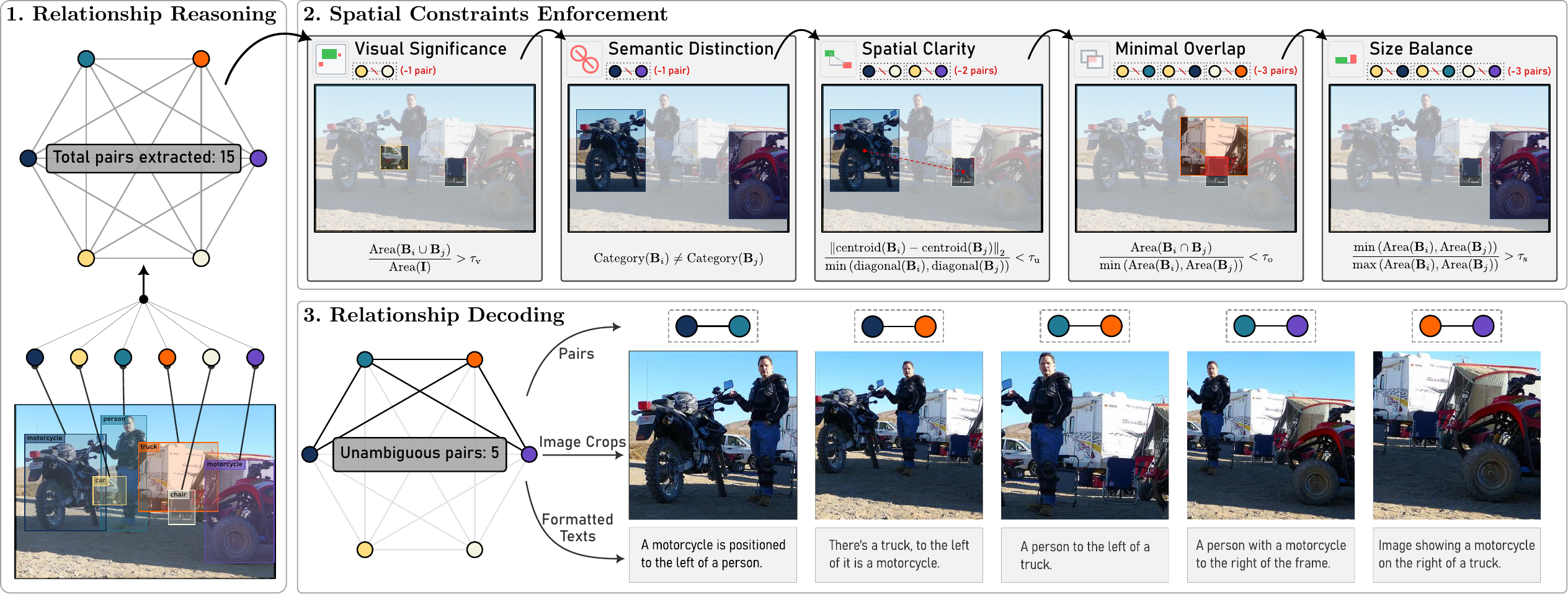}
  \caption{
    \textbf{Overview of the Spatial Constraints-Oriented Pairing
    (\dataEngineName{}) data engine.}
    \dataEngineName{} first
    \textbf{(1)} reasons about all possible object pairs in an image, then
    \textbf{(2)} validates their spatial relationships using a set of
    carefully designed constraints.  Finally, it
    \textbf{(3)} decodes the resulting unambiguous descriptors into image
    crops paired with accurate textual descriptions.
  } \label{fig@scope-overview}
\end{figure*}

We address this limitation through the Spatial Constraints-Oriented Pairing
(\dataEngineName{}) data engine, which identifies and validates spatial
relationships between object pairs through carefully designed spatial
constraints.
As illustrated in \cref{fig@scope-overview}, \dataEngineName{} processes
images through three key stages:

\begin{figure}
  \centering
  \scalebox{.9}{
    \begin{subfigure}[b]{.31\linewidth}
      \includegraphics[width=\textwidth]{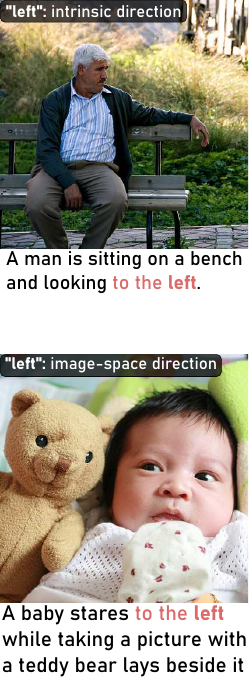}
      \caption{
        COCO~\cite{lin_microsoft_2014}
      }
    \end{subfigure}
    \begin{subfigure}[b]{.31\linewidth}
      \includegraphics[width=\textwidth]{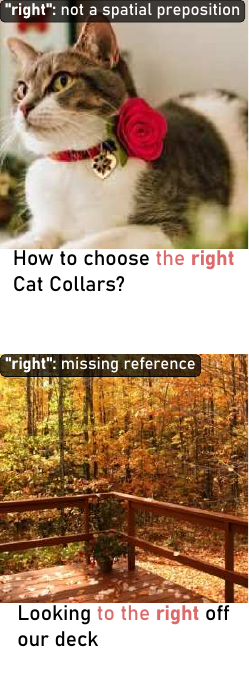}
      \caption{
        LAION~\cite{schuhmann_laion-400m_2021, schuhmann_laion-5b_2022}
      }
    \end{subfigure}
    \begin{subfigure}[b]{.31\linewidth}
      \includegraphics[width=\textwidth]{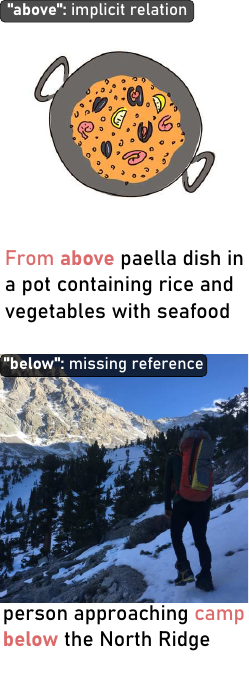}
      \caption{
        CC12M~\cite{changpinyo_conceptual_2021}
      }
    \end{subfigure}
  }
  \caption{
    \textbf{Examples highlighting common ambiguities and errors in spatial
    language annotations from COCO, LAION, and CC12M datasets.}
    \textbf{(a1, a2)} inconsistent frame of reference;
    \textbf{(b1)} non-spatial usage of spatial terms;
    \textbf{(b2, c1, c2)} missing or incorrect reference objects.
  } \label{fig@dataset-flaw}
\end{figure}

\noindent\textbf{1. Relationship Reasoning.}
The engine begins by identifying all possible pairwise relationships within
each image.
For every image $\mathbf{I}$, we first identify all valid object instances
$\{O_1, \cdots, O_n\}$, where the $i$-th object $O_i$ is associated with a
bounding box $\mathbf{B}_i$ and its corresponding category label
$\text{Category}(\mathbf{B}_i)$.
We then enumerate all possible pairs of identified objects, creating a total
of $\binom{n}{2}$ candidate pairs to be evaluated for their spatial
relationships.

\noindent\textbf{2. Spatial Constraints Enforcement.}
To ensure unambiguous spatial relationships between object pairs in images, we
formalize a set of necessary constraints that collectively guarantee visual
clarity and semantic validity.
For any candidate object pair $(O_i, O_j)$, we evaluate five constraints:

\begin{itemize}
  \item \textbf{Visual Significance:}
    Objects must occupy a sufficient portion of the image to ensure their
    spatial relationship is meaningful:
    \begin{align}
      \frac{
        \text{Area}(\mathbf{B}_i \cup \mathbf{B}_j)
      }{
        \text{Area}(\mathbf{I})
      } > \tau_\text{v}.
    \end{align}
    A relatively small value of $\tau_\text{v}$ ensures that the identified
    spatial relationship is a dominant feature of the image, rather than an
    incidental arrangement of background elements.

  \item \textbf{Semantic Distinction:}
    Objects must belong to different categories, satisfying the following
    condition:
    \begin{align}
      \text{Category}(\textbf{B}_i) \neq \text{Category}(\textbf{B}_j).
    \end{align}
    This constraint eliminates ambiguous cases where multiple instances of the
    same object category could create reference confusion in spatial
    descriptions.

  \item \textbf{Spatial Clarity:}
    Objects must be within a reasonable proximity for clear spatial reference.
    Formally, we define the distance between centroids as
    $d(i, j) = \left\lVert \text{centroid}(\mathbf{B}_i) - \text{centroid}(\mathbf{B}_j) \right\rVert_2$
    and a characteristic length as
    $l(i, j) = \min\left( \text{diagonal}(\mathbf{B}_i), \text{diagonal}(\mathbf{B}_j) \right)$.
    An object pair $(O_i, O_j)$ satisfies this constraint when:
    \begin{align}
      \frac{
        d(i, j)
      }{
        l(i, j)
      } < \tau_\text{u},
    \end{align}
    where $\tau_\text{u}$ upper-bounds the relative distance between $O_i$ and
    $O_j$, avoiding ambiguous spatial relationships between far-apart objects.

  \item \textbf{Minimal Overlap:}
    Objects must maintain sufficient visual separation:
    \begin{align}
      \frac{
        \text{Area}(\mathbf{B}_i \cap \mathbf{B}_j)
      }{
        \min\left(
          \text{Area}(\mathbf{B}_i),
          \text{Area}(\mathbf{B}_j)
        \right)
      } < \tau_\text{o}.
    \end{align}
    The overlap threshold $\tau_\text{o}$ preserves individual object
    visibility, while allowing natural spatial configurations like \vv{cup on
    table} where partial overlap is expected.

  \item \textbf{Size Balance:}
    The objects in a pair should have comparable visual prominence:
    \begin{align}
      \frac{
        \min\left(
          \text{Area}(\mathbf{B}_i),
          \text{Area}(\mathbf{B}_j)
        \right)
      }{
        \max\left(
          \text{Area}(\mathbf{B}_i),
          \text{Area}(\mathbf{B}_j)
        \right)
      } > \tau_\text{s}.
    \end{align}
    This constraint ensures that both objects contribute similarly to the
    spatial relationship, preventing cases where one object is too small to
    serve as a reliable spatial reference.
\end{itemize}
These constraints effectively filter out ambiguous spatial relationships while
preserving natural object interactions, creating a strong foundation for clear
and consistent spatial descriptions.

\noindent\textbf{3. Relationship Decoding.}
For object pairs satisfying all constraints, we adopt a representation system
that encodes the spatial relationships with structured descriptors, rather
than fixed text.
These descriptors are then decoded into image-text pairs at training time.
For example, a descriptor such as $D =$ \vv{(cup, $\mathbf{B}_\mathtt{cup}$)
<above> (couch, $\mathbf{B}_\mathtt{couch}$)} is decoded into a pair
$(\mathbf{I}_\text{roi}, \mathbf{T})$.
Here, $\mathbf{I}_\text{roi}$ is a crop of the original image containing both
$\mathbf{B}_\mathtt{cup}$ and $\mathbf{B}_\mathtt{couch}$, and $\mathbf{T}$
is a text prompt generated from a predefined pool of templates (e.g., \vv{a
cup on top of a couch} or \vv{an image of a couch below a cup}).
This decoding process captures the exact positions of the referenced objects
and their identified spatial relationship, providing both accurate textual
descriptions and flexible training signals.

\noindent\textbf{The \dataEngineName{} Dataset.} \label{par@scop-dataset}
We apply the \dataEngineName{} engine to the COCO~\cite{lin_microsoft_2014}
training split to curate a dataset for improving the spatial understanding of
T2I models.
The resulting dataset contains over \mbox{28,000} object pairs from
\mbox{15,000} images, each characterized by a well-defined spatial
relationship and an accurate textual descriptor.
The symbolic descriptors used during \dataEngineName{}'s curation process are
internal to the data engine and are not used at inference time.
The final trained model takes only standard natural language prompts as input.
The small size of this dataset, which is only a fraction of web-scale datasets
like LAION-400M (0.004\%)~\cite{schuhmann_laion-400m_2021} and CC12M
(0.13\%)~\cite{changpinyo_conceptual_2021}, highlights the data efficiency of
our targeted curation approach.
To validate the quality of our curation process, we conduct a human study
following VISOR's evaluation protocol, finding an 85.2\% agreement rate
between human annotations and our generated image-text pairs, indicating
strong alignment.

\subsection{Token Encoding Ordering Module} \label{sec@tenor}

Text-to-image diffusion models rely on text encoders to transform natural
language descriptions into semantic representations that guide the image
generation process.
To generate accurate spatial relationships, these representations must
faithfully preserve the spatial information described in the input text.
This capability is a cornerstone for effective spatial understanding in T2I
diffusion models.

\noindent\textbf{Analysis of Spatial Understanding.}
To investigate this potential failure point, we conduct a systematic analysis
of how well current text encoders represent spatial information.
Our key insight is that if text encoders were able to properly preserve
spatial relationships, then \emph{logically equivalent spatial descriptions
should yield high similarities} in their encoded representations.
To quantify this property, we design a proxy task:
given a \emph{base prompt} describing a spatial relationship (e.g., \vv{A to
the left of B}), we generate three variations of it:
(1) \emph{rephrased}: logically equivalent but differently worded (\vv{B to
the right of A}),
(2) \emph{negated relation}: substituting the relation phrase with its
opposite (\vv{A to the right of B}), and
(3) \emph{swapped entities}: exchanging object positions (\vv{B to the left of
A}).
Ideally, an encoder that understands the spatial layout should yield the
highest similarity for the \emph{rephrased} variation.

To ensure comprehensive evaluation, we conduct this analysis using all 80
object categories from COCO~\cite{lin_microsoft_2014}, with four fundamental
spatial relationships (\vv{left}, \vv{right}, \vv{above}, \vv{below}),
yielding 6,320 prompts.
For each base prompt, we identify which of the three variations yields the
most similar embedding, testing this across four different text encoders.
The results are shown in \cref{tab@text-encoder-flaw}: even
\mbox{T5-XXL}~\cite{raffel_exploring_2020} with 11B parameters fails to
recognize the logically equivalent variation \emph{over 95\% of the time}.
Lighter encoders like CLIP~\cite{dblp-clip,
dblp-cherti_reproducible_2023} show near-complete failures across different model
sizes, suggesting that current text encoders do not sufficiently preserve
spatial relationships in their encoded representations.

\begin{table}
  \caption{%
    \textbf{Most similar prompt variation retrieved by different text encoders.}
    Each encoder is used by at least one popular open-weight diffusion model 
    (SD1.5/2.1/3/XL, FLUX.1).
  }
  \label{tab@text-encoder-flaw}
  \centering
  \scalebox{0.80}{
    \begin{tabular}{l ccc c}
    \toprule
    \multirow{2}{*}{Text Encoder} & \multicolumn{3}{c}{Most Similar Variation} & \multirow{2}{*}{Correct} \\
    \cmidrule(lr){2-4}
    & Rephrased & Neg.~Rel. & Swp. \\\midrule
    CLIP ViT-L~\cite{dblp-clip} & 1 & 5088 & 1231 & 0.02\% \\
    OpenCLIP ViT-H~\cite{dblp-cherti_reproducible_2023} & 0 & 6054 & 266 & 0\% \\
    OpenCLIP ViT-bigG~\cite{dblp-cherti_reproducible_2023} & 2 & 6067 & 251 & 0.03\% \\
    T5-XXL~\cite{raffel_exploring_2020} & 306 & 4777 & 1237 & 4.84\% \\
    \bottomrule
    \end{tabular}
  }
\end{table}

\noindent\textbf{Solution.}
Based on this analysis, we propose \peName{}, a plug-and-play module designed
to address the structural ambiguity limitations of current text encoders.
\peName{} operates by augmenting the text conditioning signals with their
original token ordering information within the text-image attention layers of
the diffusion model.
Unlike standard positional encodings in transformers, which are added only
once to the initial token embeddings~\cite{vaswani_attention_2017}, \peName{}
explicitly injects this token ordering information into \emph{every}
text-image attention operation.
This design ensures that the prompt's structural and sequential information is
consistently available every time text guidance influences the image
generation process.

\begin{figure}
  \centering
  \scalebox{.93}{
    \includegraphics[width=\linewidth]{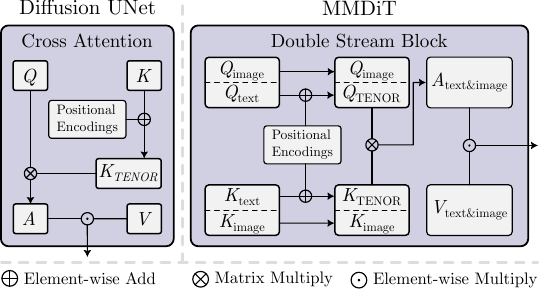}
  }
  \caption{
    \textbf{Overview of Token Encoding Ordering (\peName{}).}
    \peName{} injects token ordering information into every text-image
    attention operation in both UNet- and MMDiT-based diffusion models.
  } \label{fig@tenor-overview}
\end{figure}

\noindent\textbf{Implementation and Efficiency.}
\cref{fig@tenor-overview} illustrates the overall process of \peName{}.
Specifically, for UNet-based models~\cite{rombach_high-resolution_2022,
podell_sdxl_2023}, \peName{} adds absolute positional
encodings~\cite{vaswani_attention_2017} to the key ($K$) vectors in each
text-image attention layer.
For MMDiT-based models~\cite{peebles_scalable_2023, esser_scaling_2024}, it
adds these encodings to both the text query ($Q_\text{text}$) and key
($K_\text{text}$) vectors.
Alternative approaches for improving spatial control involve fine-tuning the
text encoder~\cite{dblp-spright}, performing test-time
optimization~\cite{chefer_attend-and-excite_2023, dblp-layout-guidance},
or incorporating large language models~\cite{dblp-dpg-bench,
wu_paragraph--image_2023}.
However, these methods typically introduce significant computational overhead
during either training or inference.
In contrast, our solution requires no additional trainable parameters and
incurs negligible computational overhead at inference time.
While adapting \peName{} requires a brief fine-tuning phase of the diffusion
model, this process is significantly more efficient than training new
spatial-aware text encoders.
Empirically, we find that this approach substantially improves spatial
understanding of both UNet-based models and the MMDiT-based FLUX.1.

It is crucial to clarify the distinct roles of \peName{} and the
\dataEngineName{} dataset.
\peName{} itself is agnostic to the semantic meaning of spatial terms like
\q{left} or \q{right}.
Its function is to provide a clear, unambiguous structural signal about the
token order that is often lost by standard text encoders (as shown in
\cref{tab@text-encoder-flaw}).
By explicitly preserving the sequence, \peName{} ensures that prompts like
\vv{A left of B} and \vv{B left of A} produce structurally different
conditioning signals for the diffusion model.
During fine-tuning, the model then learns to associate these distinct
structural signals with the correct visual outcomes provided by the
high-quality, unambiguous image-text pairs from our \dataEngineName{} dataset.
This synergy is what enables the significant improvement in spatial
understanding: \peName{} providing the structural capacity to differentiate
prompts, while \dataEngineName{} provides the semantic ground truth.

\begin{figure*}
  \centering
  \includegraphics[width=.93\textwidth]{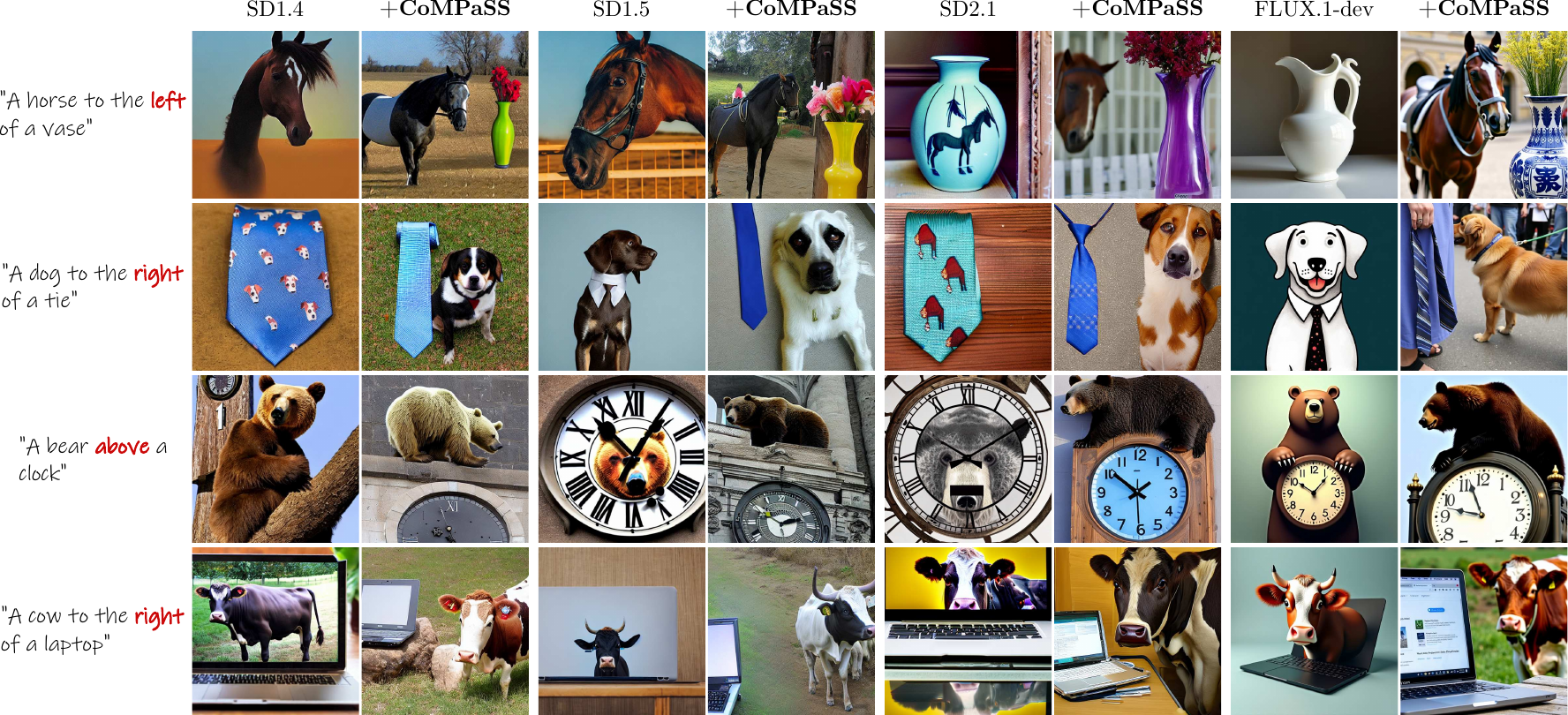}
  \caption{
    \textbf{Qualitative results of models enhanced with \frameworkName{}.}
    Our method improves the spatial understanding of both UNet-based models
    (SD1.4, SD1.5, SD2.1) and the MMDiT-based FLUX.1.
    More results are available in the supplementary material.
  } \label{fig@main-results}
\end{figure*}

\section{Experiments}

We validate the effectiveness of \frameworkName{} on four popular open-weight
text-to-image diffusion models. These models include three UNet-based
architectures (SD1.4, SD1.5, and SD2.1)~\cite{rombach_high-resolution_2022}
and one MMDiT-based architecture (FLUX.1)~\cite{BlackForestLabs2024FLUX1dev}.
Detailed experimental settings are provided in the supplementary material.

\subsection{Evaluation Details}

\begin{table*}
  \caption{
    \textbf{Comparison of base models with their \frameworkName{}-enhanced
    counterparts.}
    Our method sets a new state of the art on all spatial benchmarks,
    including VISOR~\cite{gokhale_benchmarking_2023}, T2I-CompBench
    Spatial~\cite{dblp-t2i-compbench} (T. Spat.), GenEval
    Position~\cite{dblp-geneval} (Pos.), and DPG-Bench
    Relation~\cite{dblp-dpg-bench} (Rel.).
    We also report overall (Ovr.) scores for general compositional ability on
    GenEval and DPG-Bench, and image fidelity scores using
    FID~\cite{heusel_gans_2017} and CMMD~\cite{jayasumana_rethinking_2024}.
    Detailed per-benchmark results are in the supplementary material.
  } \label{tab@metrics-main}
  \centering
  \scalebox{.83}{
    \begin{tabular}{l cccccc cc cc cc cc}
      \toprule
      \multirow{2}{*}{Method} & \multicolumn{6}{c}{VISOR (\%)} & \multicolumn{2}{c}{\makebox[0pt]{T2I-CompBench}} & \multicolumn{2}{c}{GenEval} & \multicolumn{2}{c}{DPG-Bench} & \multicolumn{2}{c}{Image Fidelity} \\
      \cmidrule(lr){2-7}
      \cmidrule(lr){8-9}
      \cmidrule(lr){10-11}
      \cmidrule(lr){12-13}
      \cmidrule(lr){14-15}
                              & uncond & cond & 1 & 2 & 3 & 4 & Spat. & Ovr. & Pos. & Ovr. & Rel. & Ovr. & FID$\downarrow$ & CMMD$\downarrow$ \\
      \midrule
      SD1.4                                     & 18.81 & 62.98 & 46.60 & 20.11 & 6.89 & 1.63 & 0.12 & 0.32 & 0.03 & 0.43 & 81.04 & 62.02 & 13.20 & 0.5618 \\
      SD1.4 +\textbf{\frameworkName{}} & \isbest{57.41} & \isbest{87.58} & \isbest{83.23} & \isbest{67.53} & \isbest{49.99} & \isbest{28.91} & \isbest{0.34} & \isbest{0.42} & \isbest{0.46} & \isbest{0.56} & \isbest{83.21} & \isbest{66.07} & \isbest{10.67} & \isbest{0.3313} \\
      \midrule
      SD1.5                            & 17.58 & 61.08 & 43.65 & 18.62 & 6.49 & 1.57 & 0.08 & 0.31 & 0.04 & 0.42 & 73.49 & 63.18 & 12.82 & 0.5548 \\
      SD1.5 +\textbf{\frameworkName{}} & \isbest{61.46} & \isbest{93.43} & \isbest{86.55} & \isbest{72.13} & \isbest{54.64} & \isbest{32.54} & \isbest{0.35} & \isbest{0.42} & \isbest{0.54} & \isbest{0.57} & \isbest{84.10} & \isbest{65.81} & \isbest{10.89} & \isbest{0.3235} \\
      \midrule
      SD2.1 & 30.25 & 63.24 & 64.42 & 35.74 & 16.13 & 4.70 & 0.13 & 0.37 & 0.07 & 0.50 & 83.95 & 65.47 & 21.65 & 0.6472 \\
      SD2.1 +SPRIGHT~\cite{dblp-spright} & \ispriorbest{43.23} & \ispriorbest{71.24} & 71.78 & \ispriorbest{51.88} & \ispriorbest{33.09} & \ispriorbest{16.15} & 0.21 & - & 0.11 & 0.51 & - & - & \isbest{16.15} & 0.512 \\
      SD2.1 +\textbf{\frameworkName{}} & \isbest{62.06} & \isbest{90.96} & \isbest{85.02} & \isbest{71.29} & \isbest{56.03} & \isbest{35.90} & \isbest{0.32} & \isbest{0.43} & \isbest{0.51} & \isbest{0.54} & \isbest{86.54} & \isbest{69.48} & 16.96 & \isbest{0.4083} \\
      \midrule
      FLUX.1 & 37.96 & 66.81 & 64.00 & 44.18 & 28.66 & 14.98 & 0.18 & 0.46 & 0.26 & 0.60 & 92.30 & 80.63 & 27.96 & 0.8737 \\
      FLUX.1 +\textbf{\frameworkName{}} & \isbest{75.17} & \isbest{93.22} & \isbest{91.73} & \isbest{83.31} & \isbest{72.21} & \isbest{53.41} & \isbest{0.30} & \isbest{0.55} & \isbest{0.60} & \isbest{0.76} & \isbest{94.12} & \isbest{84.42} & \isbest{26.40} & \isbest{0.6859} \\
      \bottomrule
    \end{tabular}
  }
\end{table*}

\begin{table}
  \caption{
    \textbf{Computational efficiency comparison.}
    Overheads \emph{exclude} bounding box generation for methods that require
    it, and are reported in the format ``Time/VRAM''.
    Lower is better.
  } \label{tab@efficiency-compute}
  \centering
  \scalebox{.87}{
    \begin{tabular}{l c c}
      \toprule
      Method & \textbf{Training} & \textbf{Inference} \\
      \midrule
      SPRIGHT~\cite{dblp-spright} & +8.1\%/+24.5\% & - \\
      R\&B~\cite{dblp-rnb} & - & +1108.7\%/+150.9\% \\
      Attn. Refocusing~\cite{dblp-attention-refocusing} & - & +23.8\%/+38.0\% \\
      Layout Guidance~\cite{dblp-layout-guidance} & - & +180.9\%/+67.4\% \\
      \textbf{\frameworkName{} (ours)}        & \textbf{+3.9\%}/\textbf{+0.7\%} & \textbf{+2.47\%}/\textbf{+0.6\%} \\
      \bottomrule
    \end{tabular}
  }
\end{table}

\begin{table}
  \caption{
    \textbf{Data efficiency comparison.}
    With a comparable amount of training data ($\sim$500 images), our method
    (\frameworkName{}$_\text{500}$) substantially outperforms prior work on
    GenEval Position (G. Pos.).
    Performance continues to scale favorably as more data is used.
    \textit{SPRIGHT was not evaluated on FLUX.1 as its methodology of
    co-training the T5-XXL text encoder exceeds our hardware resources.}
  } \label{tab@efficiency-data}
  \centering
  \scalebox{.93}{
    \begin{tabular}{l c c}
      \toprule
      Method & \# Training Images & G. Position \\
      \midrule
      SD2.1 & 0 & 0.07 \\
      SD2.1 +SPRIGHT~\cite{dblp-spright} & $\sim$500 & 0.11 \\
      SD2.1 \textbf{+\frameworkName{}}$_\text{500}$ & 500 & \textbf{0.23} \\
      SD2.1 \textbf{+\frameworkName{}} & 28k & \textbf{0.51} \\
      FLUX.1 & 0 & 0.26 \\
      FLUX.1 \textbf{+\frameworkName{}}$_\text{500}$ & 500 & \textbf{0.56} \\
      FLUX.1 \textbf{+\frameworkName{}} & 28k & \textbf{0.60} \\
      \bottomrule
    \end{tabular}
  }
\end{table}

\noindent\textbf{Baselines.}
We compare \frameworkName{} against several baselines:
1) SPRIGHT~\cite{dblp-spright} is a training-based method whose focus on
   spatial consistency is highly aligned with ours.
2) R\&B~\cite{dblp-rnb}, 3) Attention
   Refocusing~\cite{dblp-attention-refocusing}, and 4) Layout
   Guidance~\cite{dblp-layout-guidance} are inference-only methods for
   zero-shot layout control.
   These methods rely on additional image-space guidance, such as bounding
   boxes or masks, which are typically provided either through manual
   annotation or by external language models~\cite{feng_layoutgpt_2024,
   lian_llm-grounded_2023, cho_visual_2023}.

\noindent\textbf{Metrics.}
We extensively evaluate \frameworkName{} on several benchmarks with varying
complexity:
VISOR~\cite{gokhale_benchmarking_2023} is a specialized benchmark with 31,680
prompts focusing on spatial understanding.
T2I-CompBench~\cite{dblp-t2i-compbench} and GenEval~\cite{dblp-geneval} are
two benchmarks for evaluating
open-world compositional text-to-image generation.
\mbox{DPG-Bench}~\cite{dblp-dpg-bench} is a comprehensive dataset consisting
of 1,065 lengthy, dense prompts, designed to assess the intricate semantic
alignment capabilities of text-to-image models.
FID~\cite{heusel_gans_2017} and CLIP Maximum Mean Discrepancy
(CMMD)~\cite{jayasumana_rethinking_2024} assess image fidelity, with FID
focusing on photorealism and CMMD providing an alternative measure that better
aligns with human preferences.
Additionally, we measure time and VRAM overhead to evaluate the computational
efficiency of each method.

\subsection{Main Results}

\noindent\textbf{Spatial Understanding.}
We evaluate spatial relationship generation accuracy across multiple
benchmarks and report them in \cref{tab@metrics-main}.
Adding \frameworkName{} to existing diffusion models sets a new state of the
art across all spatial-related metrics.
Notably, when applied to the state-of-the-art open-weight model FLUX.1,
\frameworkName{} achieves substantial relative gains on VISOR (+98\%),
T2I-CompBench Spatial (+67\%), GenEval Position (131\%), and the best scores
on the DPG-Bench Relation benchmark.
As shown in \cref{fig@main-results}, \frameworkName{} effectively addresses
these limitations, correctly rendering novel spatial configurations unseen
during training and demonstrating a significant enhancement in spatial
understanding.

\noindent\textbf{General Generation Capability and Fidelity.}
In addition to spatial metrics, we evaluate performance on other compositional
tasks from the T2I-CompBench~\cite{dblp-t2i-compbench},
GenEval~\cite{dblp-geneval}, and DPG-Bench~\cite{dblp-dpg-bench}
benchmarks.
We also assess image fidelity using FID~\cite{heusel_gans_2017} and
CMMD~\cite{jayasumana_rethinking_2024}.
The overall results are reported in \cref{tab@metrics-main}, with detailed
per-task results provided in \cref{tab@supp-metrics-general} in the
supplementary material.
While \frameworkName{} is designed to target spatial performance, we find that
it also improves overall alignment scores and image fidelity.
We conjecture that in base models, spatial terms are often entangled with
unrelated semantics due to flawed data.
By disentangling these terms, \frameworkName{} may also help the model better
understand other aspects of language, resulting in these broader improvements.
Importantly, on DPG-Bench, which contains 1,065 long natural language prompts,
models enhanced with \frameworkName{} achieve consistently higher overall
scores, indicating that the enhanced spatial understanding generalizes well to
longer and more complex text prompts.

\noindent\textbf{Computational Efficiency.}
We compare the time and VRAM overhead of \frameworkName{} to prior work during
both training and inference, with results reported in
\cref{tab@efficiency-compute}.
SPRIGHT~\cite{dblp-spright}, a prior training-based method focused on spatial
consistency, finetunes the text encoder along with the diffusion network.
However, this approach inevitably adds computational requirements (+8.1\% in
time and +24.5\% in VRAM).
Inference-only methods~\cite{dblp-rnb, dblp-attention-refocusing,
dblp-layout-guidance} typically enable gradient-based optimization at test
time, incurring substantial overhead (\eg{} time overhead ranges from 23.8\%
up to 1108.7\%).
In contrast, \frameworkName{} only alters the cross-attention calculation and
introduces no extra trainable parameters. It is therefore lightweight,
incurring only $\sim$3\% time overhead and less than 1\% VRAM overhead during
inference.

\noindent\textbf{Data Efficiency.}
\frameworkName{} also demonstrates remarkable data efficiency.
In our experiments, we randomly sample 500 images from our 28k object pairs
dataset while preserving the distribution of categories and relationships
(\frameworkName{}$_\text{500}$, see \cref{tab@efficiency-data}).
With this comparable data volume to SPRIGHT~\cite{dblp-spright},
\frameworkName{}$_\text{500}$ achieves substantially better spatial accuracy
(0.23 vs 0.11) on SD2.1.
The performance scales favorably with more training data, reaching 0.51 when
using the full dataset.
Similarly, on the MMDiT-based FLUX.1, \frameworkName{}$_\text{500}$ improves
the spatial accuracy from 0.26 to 0.56, nearly matching the 0.60 score
achieved with the complete dataset.
This exceptional data efficiency highlights how \frameworkName{}'s design
effectively leverages the high-quality spatial relationships extracted by the
\dataEngineName{} data engine.

\subsection{Ablation Studies}

\begin{table}
  \caption{
    \textbf{Ablation study on the hyperparameters of \dataEngineName{}.}
    Our main experiments use the hyperparameter set $\{\tau_\text{v},
    \tau_\text{u}, \tau_\text{o}, \tau_\text{s} \} = \{0.2, 2.0, 0.3, 0.5\}$,
    denoted as ``\textbf{Ours}'' in the table.
    The results show that performance is robust to variations in these
    hyperparameters.
  } \label{tab@ablation-scop}
  \centering
  \scalebox{.73}{
    \begin{tabular}{l cc cc cc cc c}
      \toprule
      Name & \multicolumn{2}{c}{$\tau_\text{v}$} & \multicolumn{2}{c}{$\tau_\text{u}$} & \multicolumn{2}{c}{$\tau_\text{o}$} & \multicolumn{2}{c}{$\tau_\text{s}$} & \multirow{2}{*}{\textbf{Ours}} \\
      \cmidrule(lr){1-1}
      \cmidrule(lr){2-3}
      \cmidrule(lr){4-5}
      \cmidrule(lr){6-7}
      \cmidrule(lr){8-9}
      Value
        & 0.1 & 0.3  %
        & 1.0 & 3.0  %
        & 0.2 & 0.5  %
        & 0.3 & 0.7  %
        & \\
      \midrule
      SD1.5
        & 0.48 & 0.52
        & 0.50 & 0.49
        & 0.47 & 0.51
        & 0.51 & 0.53
        & \textbf{0.54} \\
      FLUX.1
        & 0.55 & 0.56
        & 0.59 & 0.54
        & 0.54 & 0.58
        & 0.58 & 0.59
        & \textbf{0.60} \\
      \bottomrule
    \end{tabular}
  }
\end{table}

\begin{table}
  \caption{
    \textbf{Ablation study on the components of \frameworkName{}.}
    (i) original models;
    (ii) trained with the \dataEngineName{} dataset described in
    \cref{par@scop-dataset};
    (iii) our full method.
    T2I-CompBench Spatial (T. Spatial) and GenEval Position (G. Pos) scores
    are reported.
  } \label{tab@ablation-components}
  \centering
  \scalebox{.88}{
    \begin{tabular}{cl cc cc}
      \toprule
      \multirow{2}{*}{Setting} & \multirow{2}{*}{Model} & \multicolumn{2}{c}{Components}                & \multirow{2}{*}{T. Spatial} & \multirow{2}{*}{G. Pos.} \\
      \cmidrule(lr){3-4}
      &                       & \dataEngineName{} & \peName &         &            \\
      \midrule
      (i) & SD1.5 & & & 0.08 & 0.04 \\
      (ii)& SD1.5 & \checkmark & & 0.32 & 0.39 \\
      (iii) & SD1.5 & \checkmark & \checkmark & \textbf{0.35} & \textbf{0.54} \\
      \midrule
      (i) & FLUX.1 & & & 0.18 & 0.26 \\
      (ii) & FLUX.1 & \checkmark & & 0.29 & 0.56 \\
      (iii) & FLUX.1 & \checkmark & \checkmark & \textbf{0.30} & \textbf{0.60} \\
      \bottomrule
    \end{tabular}
  }
\end{table}

\begin{figure}
  \centering
  \includegraphics[width=.93\linewidth]{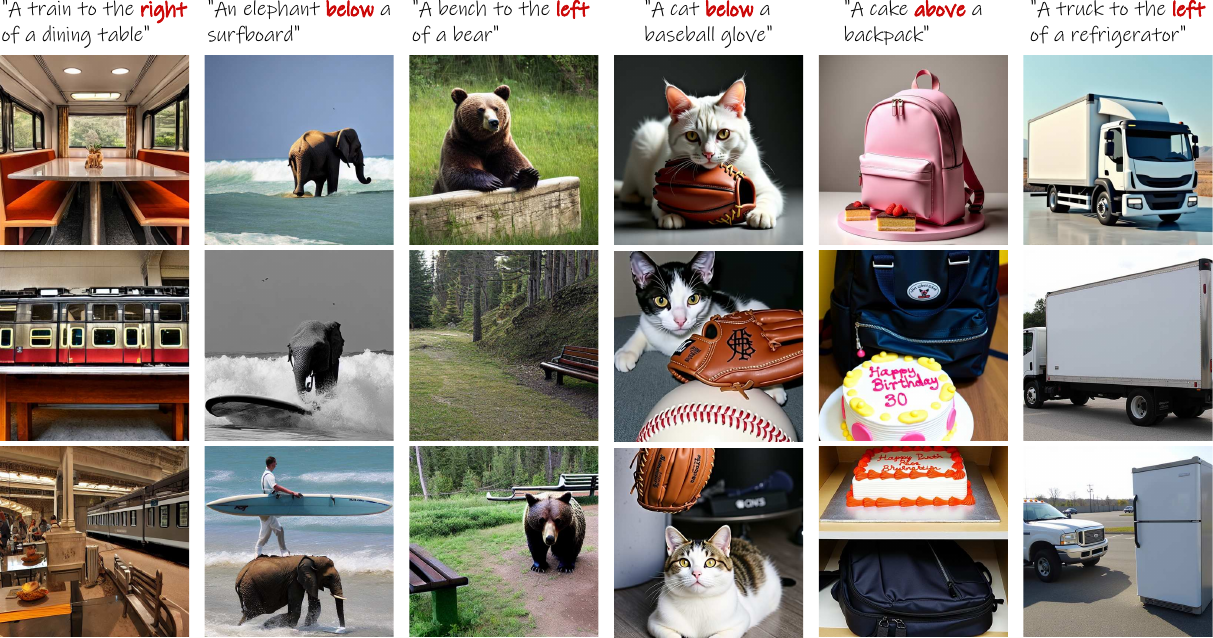}
  \caption{
    \textbf{Qualitative ablation study on the \peName{} module.}
    The \peName{} module improves generalization to unseen prompts.
    \textbf{(i)} First row: original models.
    \textbf{(ii)} Second row: models trained only with the \dataEngineName{}
    dataset.
    \textbf{(iii)} Third row: our full method.
    The left three columns show results from SD1.5; the right three columns
    show results from FLUX.1.
  } \label{fig@ablation}
\end{figure}

To understand the contribution of each component, we conduct controlled
experiments on two representative models: SD1.5, a popular UNet-based model,
and FLUX.1-dev, a state-of-the-art MMDiT-based model.

\noindent\textbf{Hyperparameters of \dataEngineName{}.}
The \dataEngineName{} data engine has four tunable hyperparameters. We
empirically determine the optimal values to be $\{\tau_\text{v}, \tau_\text{u},
\tau_\text{o}, \tau_\text{s} \} = \{0.2, 2.0, 0.3, 0.5\}$ via grid search.
In \cref{tab@ablation-scop}, we report the model's sensitivity to each of
these four hyperparameters by evaluating performance on nearby values.
While our chosen hyperparameters yield optimal results, the model's
performance remains high across a range of nearby values, indicating the
robustness of the \dataEngineName{} engine.

\noindent\textbf{Effect of \dataEngineName{}.}
In \cref{tab@ablation-components}, comparing settings (i) and (ii) within each
model reveals that \dataEngineName{} alone contributes substantially to the
spatial understanding of diffusion models.
This finding supports our initial hypothesis that a primary cause of failure
in spatial generation is the flawed nature of existing training data.
This confirms that \dataEngineName{} successfully curates high-quality data
that directly improves the spatial understanding of diffusion models.

\noindent\textbf{Effect of \peName{}.}
\cref{tab@ablation-components} shows that our full method (setting (iii)
within each model) further improves the spatial accuracy.
As shown in \cref{fig@ablation}, we evaluate the models on text descriptions
unseen during training across all three settings.
While \dataEngineName{} provides rich and accurate spatial priors, it is only
after incorporating \peName{} that the model generalizes better to challenging,
unseen spatial configurations.
This phenomenon connects closely to our initial finding about spatial
understanding deficiency in text encoders (\cref{tab@text-encoder-flaw}):
when the semantic representations of completely opposite text descriptions are
indistinguishable, the model is trained to align mixed image signals from
seemingly similar text conditions, resulting in poor generalization beyond
training data.
\peName{} addresses this issue by injecting token ordering information into
the text-image attention operations.
This ensures that structurally different prompts produce distinct conditioning
signals, enabling the model to learn correct associations from the
\dataEngineName{} data and thereby achieve better zero-shot generation on
unseen prompts.

\section{Discussion and Conclusion}

\begin{figure}[t]
  \centering
  \includegraphics[width=.93\linewidth]{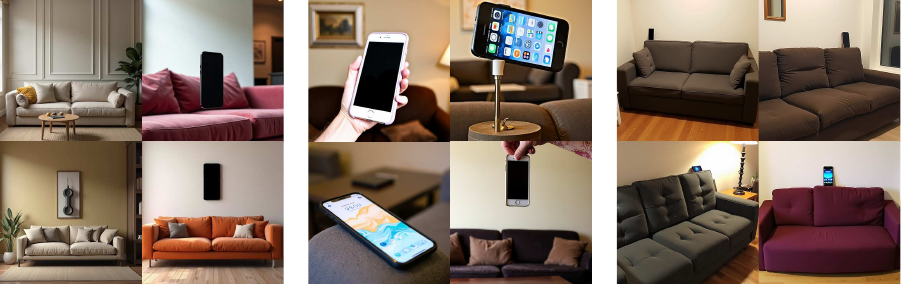}
  \caption{
    Results for \vv{a \emph{tiny} phone above a \emph{large} couch}.
    \textbf{Left}:
      FLUX.1.
    \textbf{Middle}:
      FLUX.1+CoMPaSS.
    \textbf{Right}:
      Fine-tuning (100 steps, batch size 4) on FLUX.1+CoMPaSS with only
      size-contrast data filtered from adapted \dataEngineName{}.
  } \label{fig@size-contrast}
\end{figure}

\noindent\textbf{Limitations and Future Work.}
While \frameworkName{} demonstrates robust performance, its understanding of
spatial language is not exhaustive.
Our framework primarily addresses ambiguity arising from noisy web data and
encoder limitations, but does not yet model the rich, context-dependent nature
of spatial language, a key topic in Qualitative Spatial Relations
(QSR)~\cite{qsrsurvey, alomari2017natural, coventry2001interplay}.
For instance, rendering subjects with extreme size disparities remains a
challenge (see \cref{fig@size-contrast}, left and middle), as the
interpretation of \vv{above} or \vv{beside} depends heavily on object context.
Our preliminary experiments show that a targeted fine-tuning stage can
mitigate this specific issue (see \cref{fig@size-contrast}, right).
Furthermore, our current approach is viewer-centric and does not handle
object-centric frames (e.g., \vv{to the child's right hand side}).
Fully integrating QSR principles and modeling both object-centric and
context-dependent relations are important directions for future work.

\noindent\textbf{Conclusion.}
We have presented \frameworkName{}, a versatile framework that enhances
spatial understanding in T2I models.
\frameworkName{} uses the \dataEngineName{} engine to curate high-quality,
spatially-unambiguous training data, and the \peName{} module to preserve
crucial token ordering information that text encoders often lose.
Extensive experiments show that \frameworkName{} sets a new state of the art
on key spatial benchmarks, achieving substantial relative gains on VISOR
(+98\%), T2I-CompBench Spatial (+67\%), and GenEval Position (+131\%) without
compromising general generation capabilities or image quality.
By addressing a fundamental limitation in T2I models, \frameworkName{}
represents an important step towards reliable, spatially-aware image
generation.

\section*{Acknowledgements}

This work was supported by the National Natural Science Foundation of China
(NSFC) under Grant No.~62032011 and the State Key Laboratory of CAD\&CG at
Zhejiang University.

{
    \small
    \bibliographystyle{ieeenat_fullname}
    \bibliography{compass}
}
\clearpage
\section*{Appendix}
\setcounter{page}{1}
\renewcommand{\thetable}{A\arabic{table}}
\renewcommand*{\thefigure}{A\arabic{figure}}
\renewcommand{\theequation}{A\arabic{equation}}
\appendix
\section{Additional Training Details} \label{sec@supp-training-details}

\subsection{\dataEngineName{} Dataset Construction} \label{sec@supp-dataset-construction}

The optimal threshold values for the \dataEngineName{} data engine were
empirically determined to be $\{ \tau_\text{v}, \tau_\text{u}, \tau_\text{o},
\tau_\text{s} \} = \{0.2, 2.0, 0.3, 0.5\}$ through grid search.
We then applied these thresholds to process the COCO training
split~\cite{lin_microsoft_2014}, resulting in the curated \dataEngineName{}
dataset.
The initial Relationship Reasoning stage identifies 2,468,858 object pairs
with their corresponding spatial relationships.
The subsequent Spatial Constraints Enforcement stage systematically filters
these pairs. This process applies a series of increasingly stringent criteria:
Visual Significance eliminates 1,929,560 pairs, Semantic Distinction removes
169,973, Spatial Clarity filters out 119,457, Minimal Overlap excludes 148,376,
and Size Balance removes a final 73,464 pairs.
This cascade ultimately yields 28,028 clear and unambiguous object pairs.

During the training phase, the spatial relationships between object pairs
are expressed with one of 8 distinct relationship tokens: \texttt{<left>},
\texttt{<above>}, \texttt{<right>}, \texttt{<below>}, \texttt{<left+above>},
\texttt{<right+above>}, \texttt{<left+below>}, and \texttt{<right+below>}.
These relationships are determined based on the exact positions of the objects,
with occasional random replacements using the \texttt{<and>} token to enhance
robustness.
For each pair, a square region containing both objects is automatically chosen,
which is then randomly perturbed and expanded by up to 10\% before being
cropped to create the final training image paired with its corresponding text
description.

\subsection{Model Setups} \label{sec@sec-model-setups}
\begin{table*}
  \caption{
    \textbf{Hyperparameters used during training.}
  } \label{tab@supp-hyperparameters}
  \centering
  \scalebox{.985}{
    \begin{tabular}{l cccc}
      \toprule
      \textbf{Hyperparameter} & \textbf{SD1.4} & \textbf{SD1.5} & \textbf{SD2.1} & \textbf{FLUX.1} \\
      \midrule
      AdamW Learning Rate (LR) & 5e-6 & 5e-6 & 5e-6 & 1e-4 \\
      AdamW $\beta_1$ & 0.9 & 0.9 & 0.9 & 0.9 \\
      AdamW $\beta_2$ & 0.999 & 0.999 & 0.999 & 0.999 \\
      AdamW $\epsilon$ & 1e-8 & 1e-8 & 1e-8 & 1e-8 \\
      AdamW Weight Decay & 1e-2 & 1e-2 & 1e-2 & 1e-2 \\
      \midrule
      LR scheduler & Constant & Constant & Constant & Constant \\
      LR warmup steps & 0 & 0 & 0 & 20 \\
      Training Steps & 24,000 & 24,000 & 80,000 & 24,000 \\
      Local Batch Size & 1 & 1 & 1 & 1 \\
      Gradient Accumulation & 2 & 2 & 2 & 2 \\
      Training GPUs & 2$\times$L40S & 2$\times$L40S & 2$\times$L40S & 2$\times$L40S \\
      Training Resolution & $512\times512$ & $512\times512$ & $512\times512$ & $512\times512$ \\
      Trained Parameters & \multicolumn{3}{c}{All parameters of diffusion UNet} & LoRA (rank=16) on all DoubleStreamBlocks \\
      Prompt Dropout Probability & 10\% & 10\% & 10\% & 10\% \\
      \bottomrule
    \end{tabular}
  }
\end{table*}

Our experiments are conducted on four diffusion models, including three
UNet-based diffusion models
SD1.4\footnote{\url{https://huggingface.co/CompVis/stable-diffusion-v1-4}},
SD1.5\footnote{\url{https://huggingface.co/stable-diffusion-v1-5/stable-diffusion-v1-5}},
SD2.1\footnote{\url{https://huggingface.co/stabilityai/stable-diffusion-2-1}},
and the state-of-the-art MMDiT-based diffusion model
FLUX.1-dev\footnote{\url{https://huggingface.co/black-forest-labs/FLUX.1-dev}}.

For the UNet-based models, we found that incorporating attention
supervision as proposed by \cite{wang_tokencompose_2024} significantly
enhanced the convergence of the \peName{} module.  We therefore integrated this
supervision across all UNet-based implementations.
In contrast, for the FLUX.1 model, the standard denoising loss alone was
sufficient to achieve optimal performance.
Notably, our best results were achieved with the \frameworkName{}-enhanced
FLUX.1 model, using a rank-16 LoRA checkpoint that requires only $\sim$50MiB
on-disk storage, making it highly efficient for practical applications.

Detailed training hyperparameters for all model configurations are provided in
\cref{tab@supp-hyperparameters}.

\section{Runtime Performance of \peName{}} \label{sec@supp-performance}

The \peName{} module represents the only potential source of additional
computational overhead in our framework \frameworkName{}, as it injects token
ordering information into each text-image attention operation within the
diffusion models.
To quantify its impact, we conducted comprehensive benchmarking of inference
latency across all model configurations:

\begin{center}
  \scalebox{.93}{
    \begin{tabular}{l c c}
      \toprule
      \textbf{Model} & \textbf{Latency} @ $512\times512$ & \textbf{Overhead} \\
      \midrule
      SD1.4             & 1.17s $\pm$ 3.04ms & \multirow{2}{*}{+0.85\%} \\
      SD1.4 +\peName{}  & 1.18s $\pm$ 7.24ms & \\
      \midrule
      SD1.5             & 1.17s $\pm$ 2.50ms & \multirow{2}{*}{+1.71\%} \\
      SD1.5 +\peName{}  & 1.19s $\pm$ 4.70ms & \\
      \midrule
      SD2.1             & 1.13s $\pm$ 2.50ms & \multirow{2}{*}{+4.42\%} \\
      SD2.1 +\peName{}  & 1.18s $\pm$ 4.70ms & \\
      \midrule
      FLUX.1            & 17.3s $\pm$ 40.6ms & \multirow{2}{*}{+2.89\%} \\
      FLUX.1 +\peName{} & 17.8s $\pm$ 88.8ms & \\
      \bottomrule
    \end{tabular}
  }
\end{center}

Our measurements demonstrate that the \peName{} module has minimal impact on
runtime performance, introducing only negligible computational overhead.
Even in the most demanding case of the FLUX.1-dev model, the additional time
penalty amounts to just 2.89\% of the total inference time, making it a highly
practical enhancement for real-world applications.

\section{Additional Results} \label{sec@supp-more-experimental-results}

\subsection{Visualization of \dataEngineName{} Data}
\label{sec@supp-scop-yielded}

\begin{figure*}
  \centering
  \includegraphics[width=.83\linewidth]{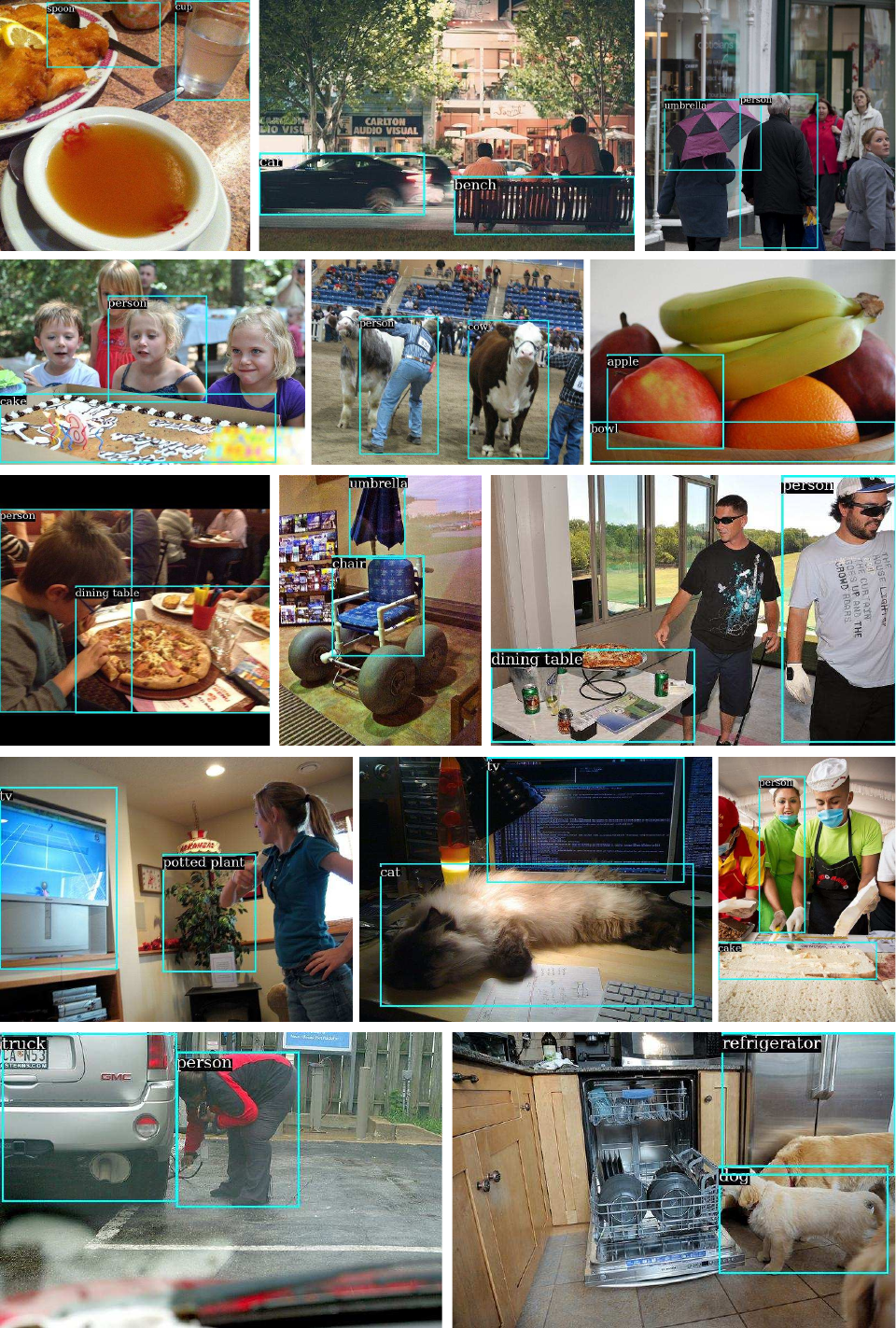}
  \caption{
    \textbf{Example object pairs and their corresponding bounding boxes
    extracted by the \dataEngineName{} data engine.}
    Each pair satisfies our spatial constraints for Visual Significance,
    Semantic Distinction, Spatial Clarity, Minimal Overlap, and Size Balance,
    ensuring unambiguous spatial relationships.
  } \label{fig@supp-scop-yielded}
\end{figure*}

We provide visualizations of data curated by \dataEngineName{} in
\cref{fig@supp-scop-yielded}.

\subsection{Additional Comparisons on VISOR} \label{sec@supp-full-visor}

We present a comprehensive comparison of VISOR metrics against other
state-of-the-art models in \cref{tab@supp-full-visor}, demonstrating the
superior performance of our approach across various evaluation criteria.

\subsection{More Visual Comparisons} \label{sec@supp-visual-resuts}
\begin{figure*}
  \centering
  \includegraphics[width=.88\linewidth]{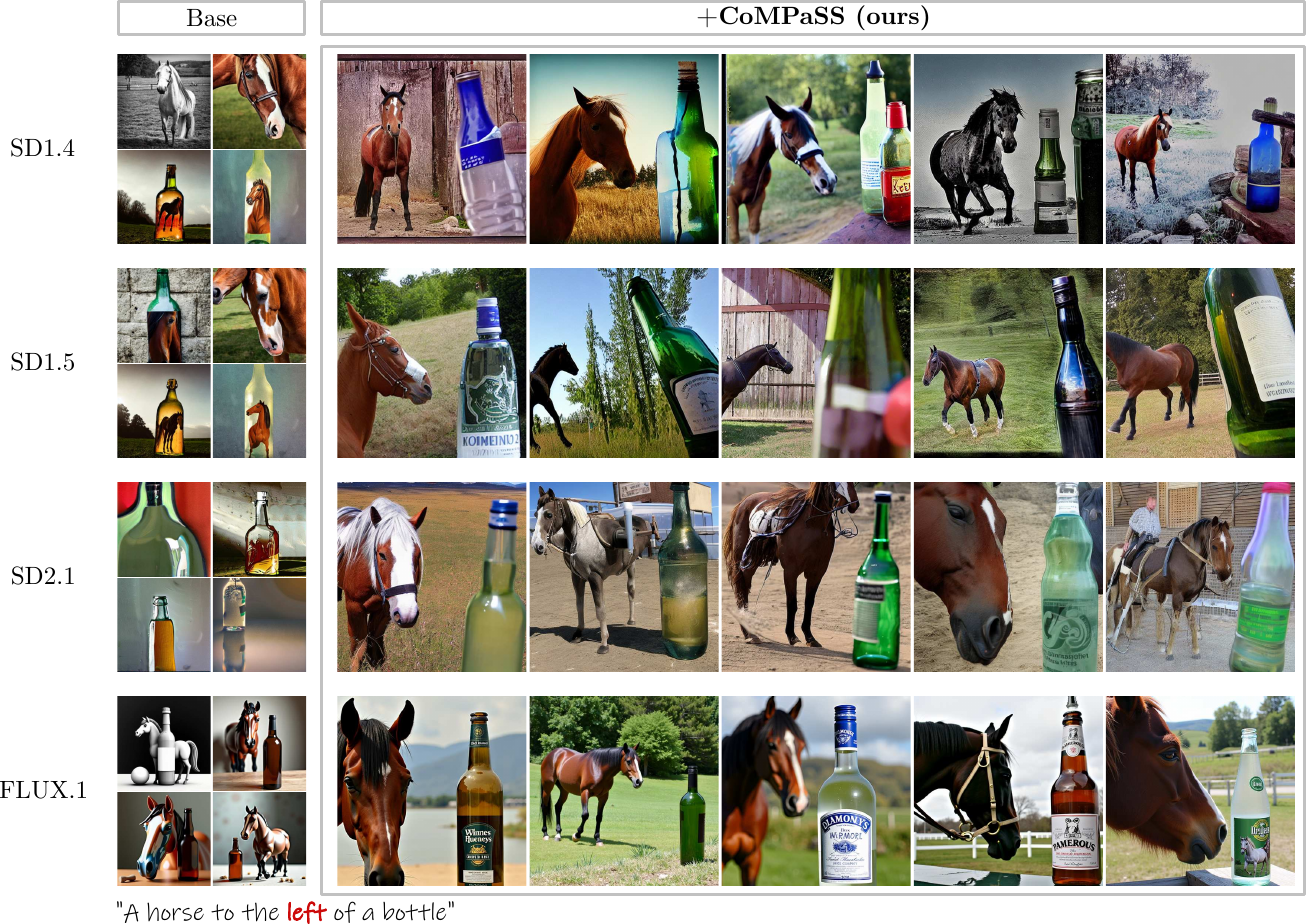}
  \includegraphics[width=.88\linewidth]{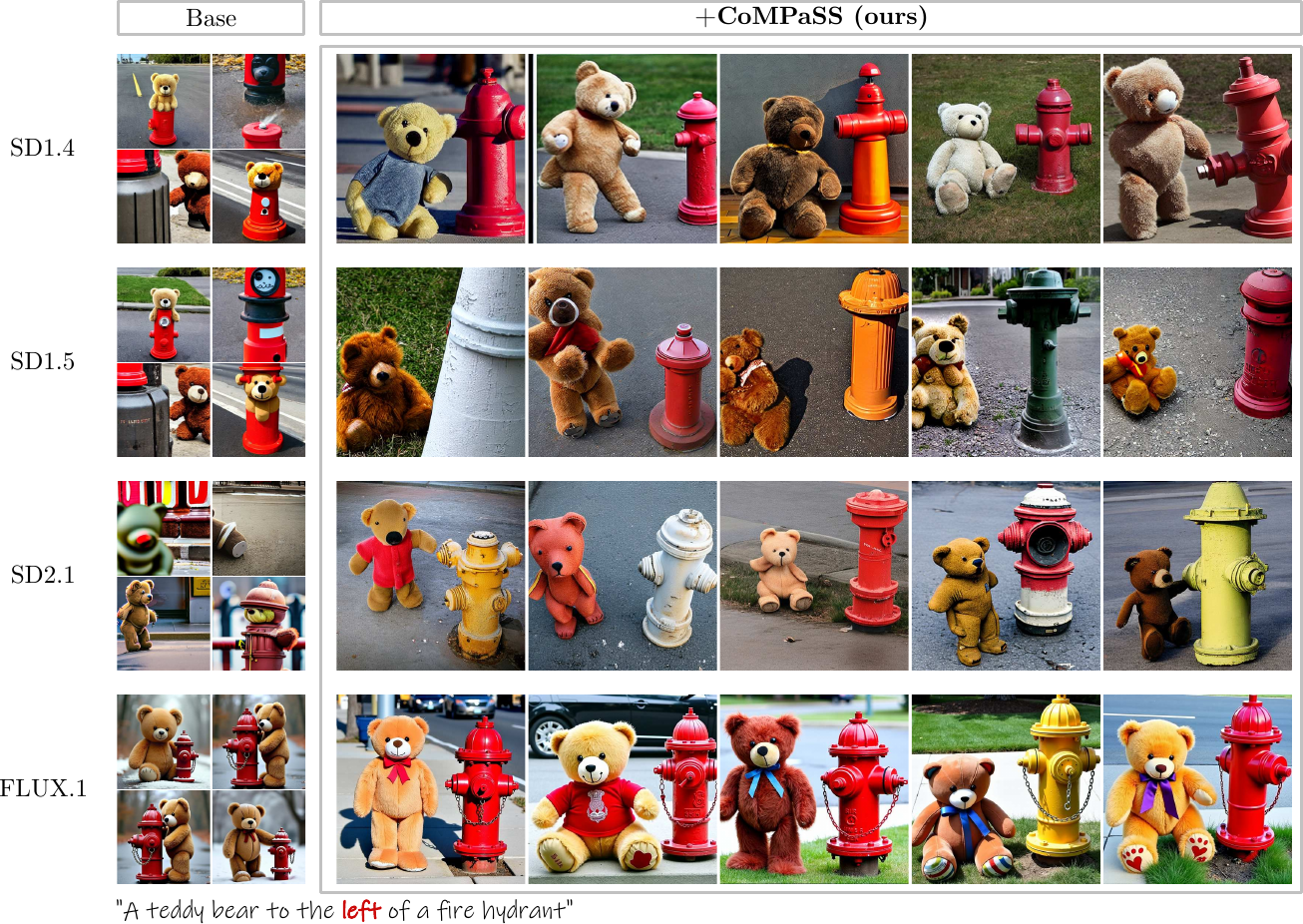}
  \caption{
    \textbf{Additional results demonstrating spatial relationship \vv{left}.}
    Our method consistently improves spatial accuracy over the baseline models.
  } \label{fig@supp-visual-left}
\end{figure*}

\begin{figure*}
  \centering
  \includegraphics[width=.88\linewidth]{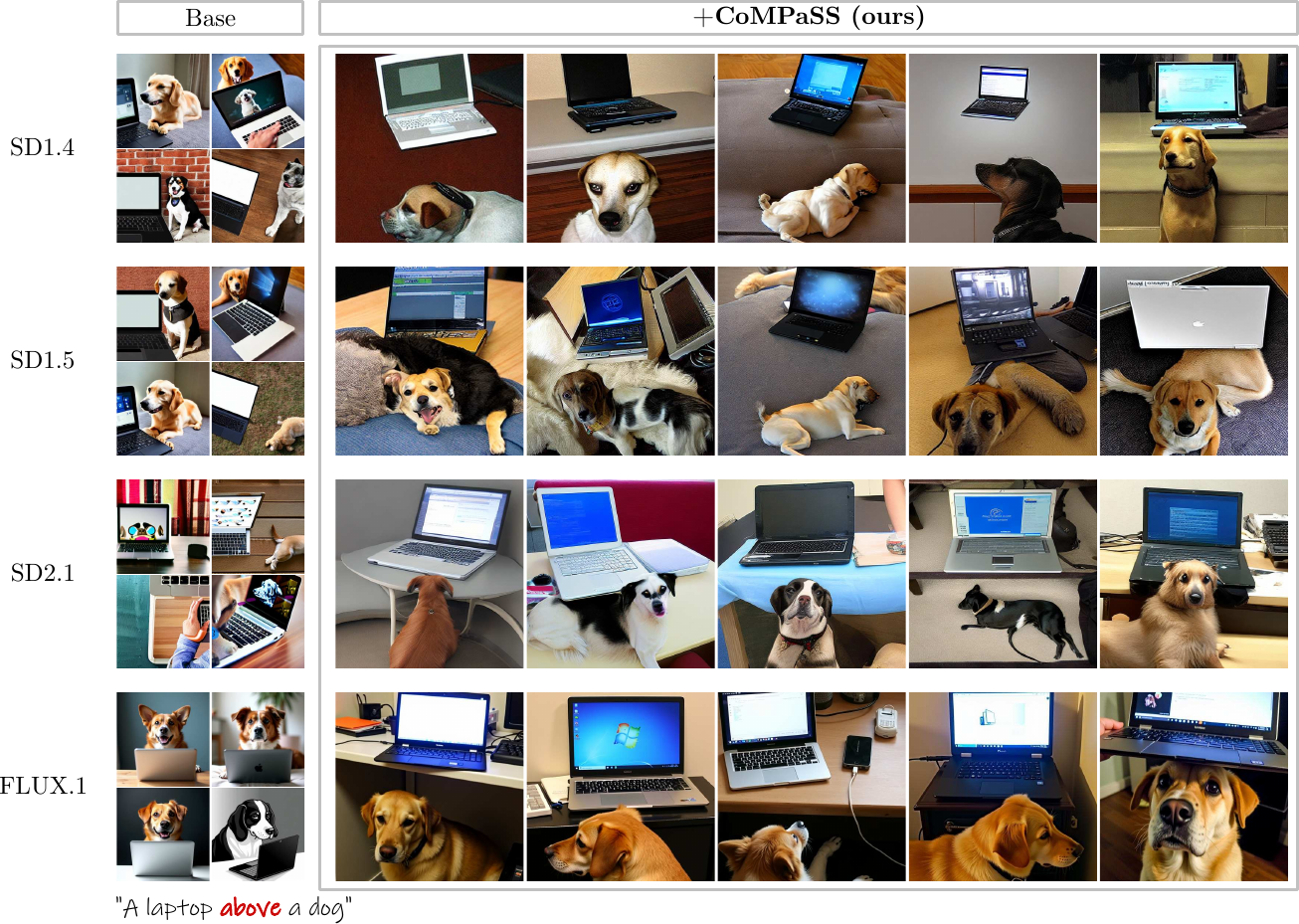}
  \includegraphics[width=.88\linewidth]{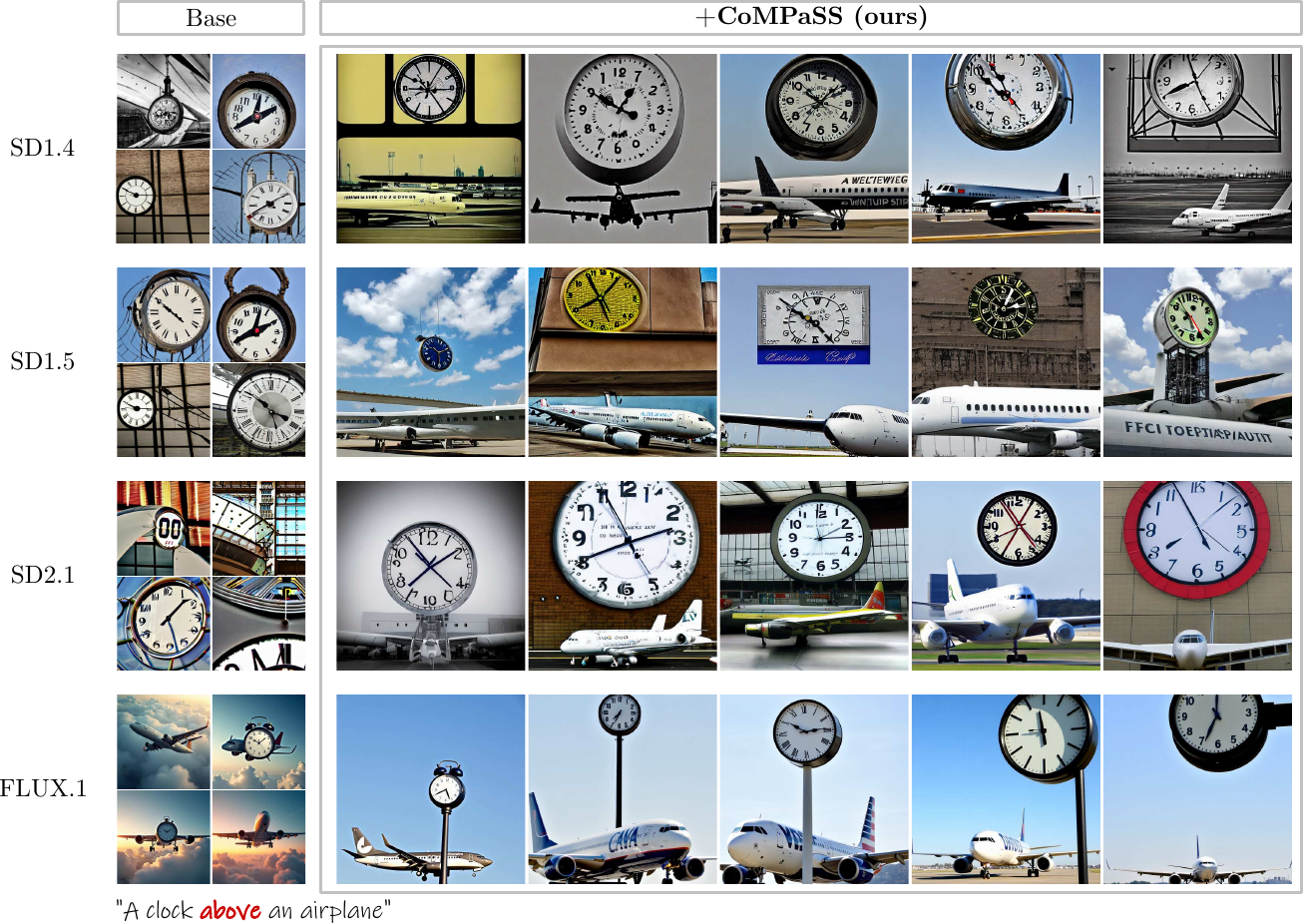}
  \caption{
    \textbf{Additional results demonstrating spatial relationship \vv{above}.}
    Our method consistently improves spatial accuracy over the baseline models.
  } \label{fig@supp-visual-above}
\end{figure*}

\begin{figure*}
  \centering
  \includegraphics[width=.88\linewidth]{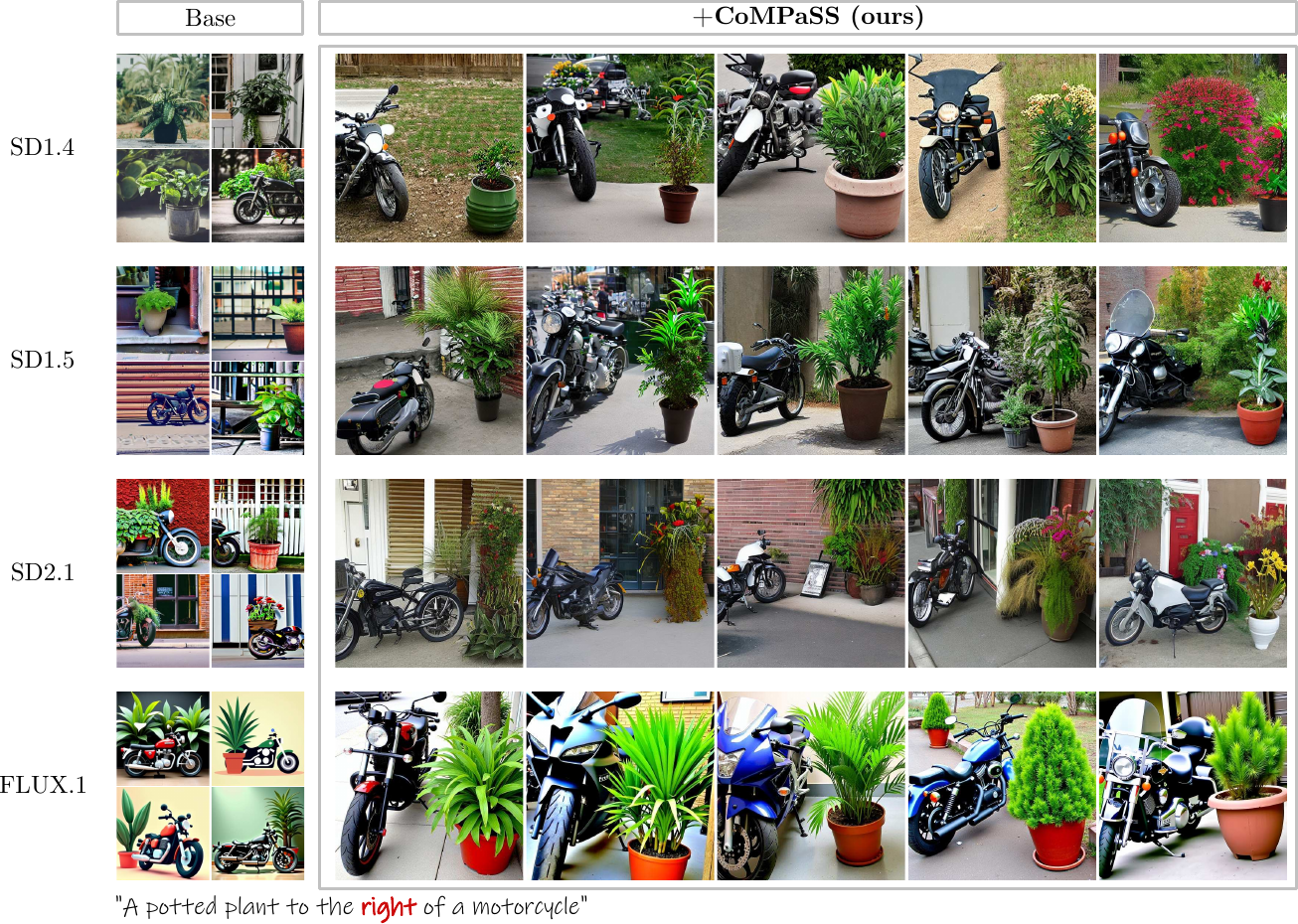}
  \includegraphics[width=.88\linewidth]{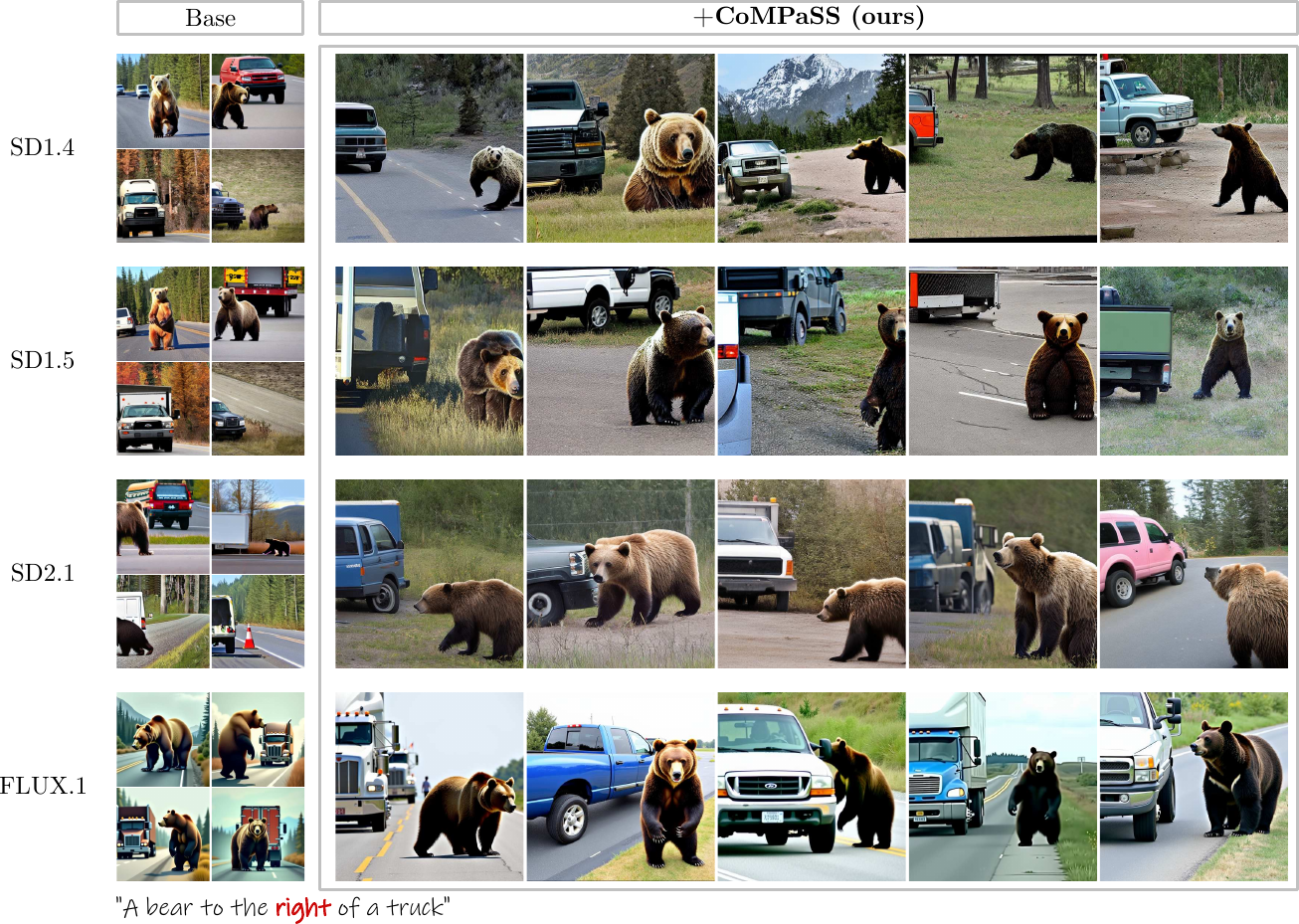}
  \caption{
    \textbf{Additional results demonstrating spatial relationship \vv{right}.}
    Our method consistently improves spatial accuracy over the baseline models.
  } \label{fig@supp-visual-right}
\end{figure*}

\begin{figure*}
  \centering
  \includegraphics[width=.88\linewidth]{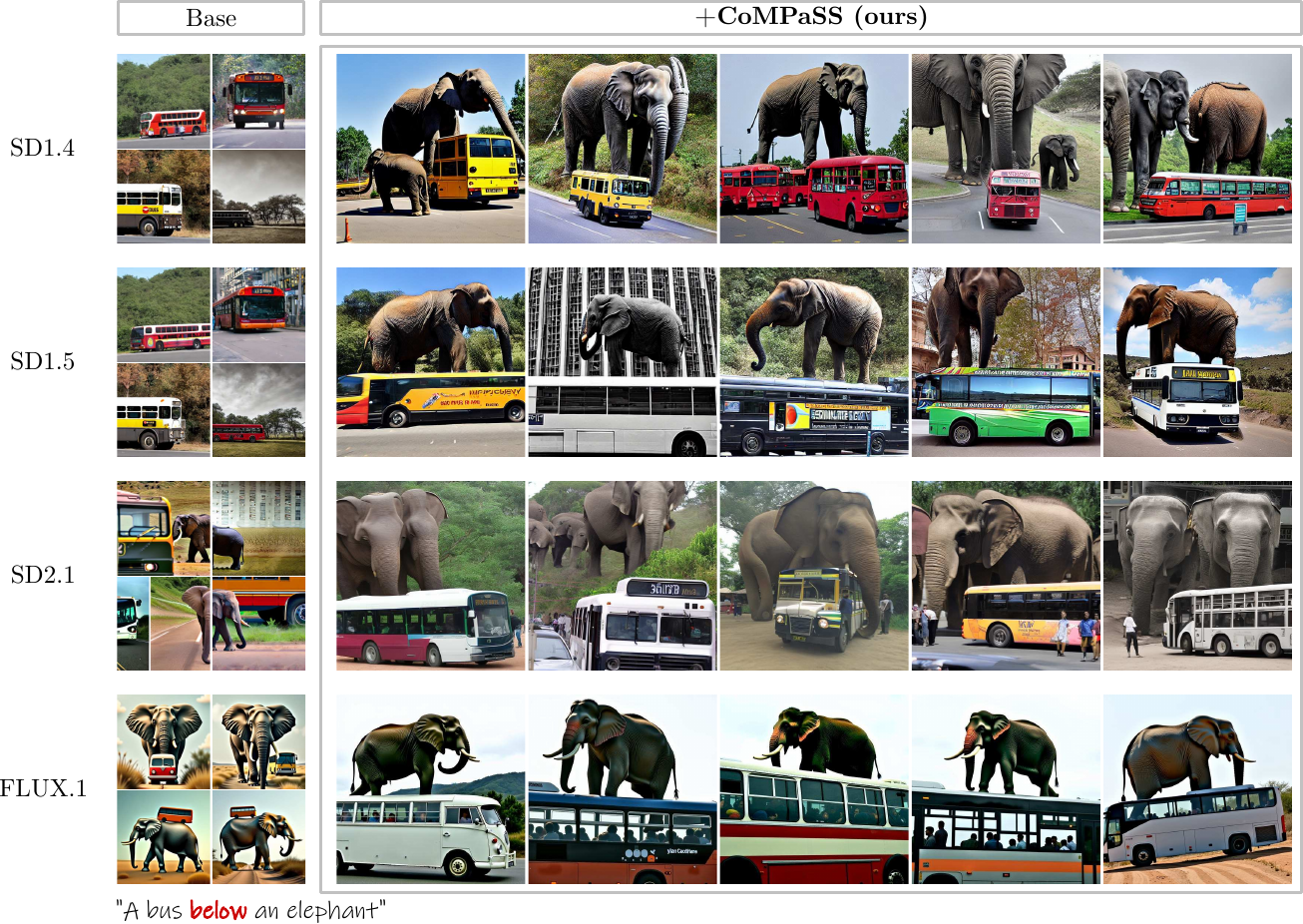}
  \includegraphics[width=.88\linewidth]{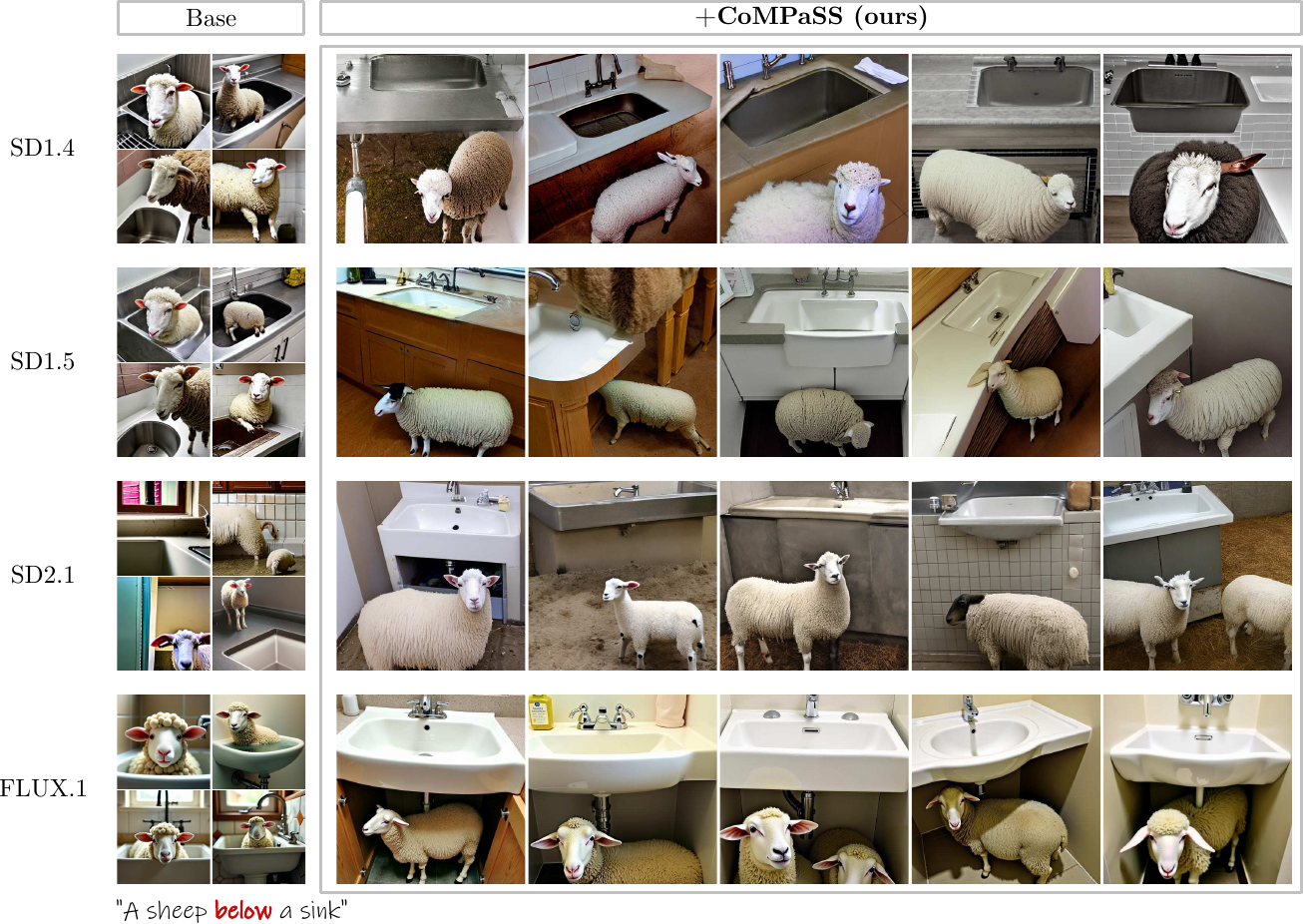}
  \caption{
    \textbf{Additional results demonstrating spatial relationship \vv{below}.}
    Our method consistently improves spatial accuracy over the baseline models.
  } \label{fig@supp-visual-below}
\end{figure*}

To provide a clear visualization of our approach's effectiveness, we present
additional visual comparisons across eight spatial configurations in
\cref{fig@supp-visual-above,fig@supp-visual-left,fig@supp-visual-below,fig@supp-visual-right}.
For each prompt, we generate a total of 36 images using different model
configurations, offering a comprehensive view of how our method consistently
improves spatial understanding across various scenarios and model
architectures.

\subsection{Evaluation on Three-Object Compositions}

To further test the generalization capabilities of \frameworkName{}, we
constructed a more challenging benchmark focused on three-object spatial
configurations. This benchmark contains 512 prompts, such as \vv{a clock above
a broccoli, with a car to the left of the clock}, and is evaluated using an
extension of the GenEval methodology~\cite{dblp-geneval}. Despite not being
fine-tuned on such complex prompts, \frameworkName{} demonstrates a
significant improvement in accuracy, as detailed in
\cref{tab@multi-object-eval}. This result builds upon the strong two-object
performance shown in the main paper (\cref{tab@metrics-main}).

\begin{table}
  \caption{
    \textbf{Multi-object spatial relationship evaluation.}
    Despite being trained only on object pairs, \frameworkName{} improves the
    model's spatial accuracy on prompts involving three objects.
  } \label{tab@multi-object-eval}
  \centering
  \scalebox{.93}{
  \begin{tabular}{l cc}
    \toprule
    Model & \% Correct (any) & \% Correct (all) \\
    \midrule
    SD2.1 & 9.1 & 3.3 \\
    SD2.1 \textbf{+\frameworkName{}} & \textbf{32.3} & \textbf{20.8} \\
    FLUX.1 & 30.12 & 13.1 \\
    FLUX.1 \textbf{+\frameworkName{}} & \textbf{52.44} & \textbf{31.2} \\
    \bottomrule
  \end{tabular}
  }
\end{table}

\subsection{Detailed Benchmark Breakdown} 

While the main paper reported overall scores for standard benchmarks
(\cref{tab@metrics-main}), here we provide a more granular breakdown.
\Cref{tab@supp-metrics-general} presents the full, per-task results for
T2I-CompBench, GenEval~\cite{dblp-geneval}, and
DPG-Bench~\cite{dblp-dpg-bench}, offering a comprehensive performance
overview.

\subsection{Ablation Studies on other Models}

While \cref{tab@ablation-components} in the main paper presented ablation
studies on SD1.5 and FLUX.1, here we extend our analysis to the other two
UNet-based diffusion models SD1.4 and SD2.1 in \cref{tab@supp-ablation-full}.
The results consistently demonstrate that both components of \frameworkName{}
contribute to improved spatial understanding across different model
architectures.

\begin{table}
  \caption{
    \textbf{Ablation study on the components of \frameworkName{}.}
    (i) original models;
    (ii) trained with the \dataEngineName{} dataset described in
    Sec.~3.1 of the main paper;
    (iii) our full method.
    T2I-CompBench Spatial (T. Spatial) and GenEval Position (G. Pos) scores
    are reported.
  } \label{tab@supp-ablation-full}
  \centering
  \scalebox{.92}{
    \begin{tabular}{cl cc cc}
      \toprule
      \multirow{2}{*}{Setting} & \multirow{2}{*}{Model} & \multicolumn{2}{c}{Components}                & \multirow{2}{*}{T. Spatial} & \multirow{2}{*}{G. Pos.} \\
      \cmidrule(lr){3-4}
      &                       & \dataEngineName{} & \peName &         &            \\
      \midrule
      (i) & SD1.4 & & & 0.12 & 0.03 \\
      (ii)& SD1.4 & \checkmark & & 0.29 & 0.36 \\
      (iii) & SD1.4 & \checkmark & \checkmark & \textbf{0.34} & \textbf{0.46} \\
      \midrule
      (i) & SD2.1 & & & 0.13 & 0.07 \\
      (ii) & SD2.1 & \checkmark & & 0.30 & 0.36 \\
      (iii) & SD2.1 & \checkmark & \checkmark & \textbf{0.32} & \textbf{0.51} \\
      \bottomrule
    \end{tabular}
  }
\end{table}

\begin{table*}
  \caption{
    \textbf{Comparison to state-of-the-art models on the
    VISOR~\cite{gokhale_benchmarking_2023} benchmark.}
    OA stands for \q{object accuracy}, which measures the rate at which all
    prompted objects appear in the generated image.
  } \label{tab@supp-full-visor}
  \centering
  \scalebox{.985}{
    \begin{tabular}{l cc cccc c}
      \toprule
      Method & uncond & cond & 1 & 2 & 3 & 4 & OA \\
      \midrule
      GLIDE~\cite{nichol_glide_2022}                      &  1.98 & 59.06 &  6.72 &  1.02 & 0.17 & 0.03 &  3.36 \\
      DALLE-mini~\cite{Dayma_DALL·E_Mini_2021}            & 16.17 & 59.67 & 38.31 & 17.50 & 6.89 & 1.96 & 27.10 \\
      CogView2~\cite{ding_cogview2_2022}                  & 12.17 & 65.89 & 33.47 & 11.43 & 3.22 & 0.57 & 18.47 \\
      Structured Diffusion~\cite{feng_training-free_2023} & 17.87 & 62.36 & 44.70 & 18.73 & 6.57 & 1.46 & 28.65 \\
      DALLE-2~\cite{ramesh_hierarchical_2022}   & 37.89 & 59.27 & \ispriorbest{73.59} & 47.23 & 23.26 & 7.49 & \ispriorbest{63.93} \\

      SD1.4                            & 18.81 & 62.98 & 46.60 & 20.11 & 6.89 & 1.63 & 29.86 \\
      SD1.5                            & 17.58 & 61.08 & 43.65 & 18.62 & 6.49 & 1.57 & 28.79 \\
      SD2.1 & 30.25 & 63.24 & 64.42 & 35.74 & 16.13 & 4.70 & 47.83 \\
      SD2.1 +SPRIGHT~\cite{dblp-spright} & \ispriorbest{43.23} & \ispriorbest{71.24} & 71.78 & \ispriorbest{51.88} & \ispriorbest{33.09} & \ispriorbest{16.15} & 60.68 \\
      FLUX.1 & 37.96 & 66.81 & 64.00 & 44.18 & 28.66 & 14.98 & 56.95 \\
      SD1.4 +\textbf{\frameworkName{}} & {57.41} & {87.58} & {83.23} & {67.53} & {49.99} & {28.91} & 65.56 \\
      SD1.5 +\textbf{\frameworkName{}} & {61.46} & \isbest{93.43} & {86.55} & {72.13} & {54.64} & {32.54} & 65.78 \\
      SD2.1 +\textbf{\frameworkName{}} & {62.06} & {90.96} & {85.02} & {71.29} & {56.03} & {35.90} & 68.23 \\
      FLUX.1 +\textbf{\frameworkName{}} & \isbest{75.17} & {93.22} & \isbest{91.73} & \isbest{83.31} & \isbest{72.21} & \isbest{53.41} & \isbest{78.64} \\
      \bottomrule
    \end{tabular}
  }
\end{table*}

\begin{table*}
  \caption{
    \textbf{Evaluation results of general generation capabilities across
    a wide range of tasks on GenEval~\cite{dblp-geneval},
    T2I-CompBench~\cite{dblp-t2i-compbench}, and
    DPG-Bench~\cite{dblp-dpg-bench}.}
    While designed to target spatial performance, \frameworkName{} also
    improves overall (Ovr.) alignment scores across most tasks.
  } \label{tab@supp-metrics-general}
  \centering
  \scalebox{.72}{
    \begin{tabular}{l c cccc c cccccc c ccccc}
      \toprule
      \multirow{2}{*}{Method} & \multicolumn{5}{c}{T2I-CompBench} & \multicolumn{7}{c}{GenEval} & \multicolumn{6}{c}{DPG-Bench} \\
      \cmidrule(lr){2-6}
      \cmidrule(lr){7-13}
      \cmidrule(lr){14-19}
      & \textbf{Spat.} & Col. & Shp. & Tex. & N.Sp.
        & \textbf{Pos.} & S.O. & T.O. & Count & Col. & Attr. & \textbf{Ovr.}
        & \textbf{Rel.} & Glb. & Ent. & Attr. & Other & \textbf{Ovr.}
        \\
      \midrule
      SD1.4      & 0.12 & 0.38 & 0.36 & 0.42 & 0.31
        & 0.03 & 0.98 & 0.41 & 0.34 & 0.74 & 0.06 & 0.43
        & 81.04 & 78.12 & 72.23 & 72.56 & 59.60 & 62.02 \\
      SD1.4 +\textbf{\frameworkName{}} & \isbest{0.34} & 0.49 & 0.43 & 0.53 & 0.31
        & \isbest{0.46} & 0.99 & 0.68 & 0.34 & 0.73 & 0.17 & \isbest{0.56}
        & \isbest{83.21} & 79.33 & 75.33 & 72.09 & 68.00 & \isbest{66.07} \\
      \midrule
      SD1.5 & 0.08 & 0.38 & 0.37 & 0.42 & 0.31
        & 0.04 & 0.96 & 0.38 & 0.36 & 0.75 & 0.06 & 0.42
        & 73.49 & 74.63 & 74.23 & 75.39 & 67.81 & 63.18 \\
      SD1.5 +\textbf{\frameworkName{}} & \isbest{0.35} & 0.50 & 0.43 & 0.52 & 0.31
        & \isbest{0.54} & 0.99 & 0.69 & 0.34 & 0.72 & 0.15 & \isbest{0.57}
        & \isbest{84.10} & 82.67 & 75.20 & 73.58 & 60.80 & \isbest{65.81} \\
      \midrule
      SD2.1 & 0.13 & 0.51 & 0.42 & 0.49 & 0.31
        & 0.07 & 0.98 & 0.51 & 0.44 & 0.85 & 0.17 & 0.50
        & 83.95 & 81.16 & 74.47 & 75.29 & 53.60 & 65.47 \\
      SD2.1 +SPRIGHT~\cite{dblp-spright} & 0.21 & - & - & - & -
        & 0.11 & 0.99 & 0.59 & 0.49 & 0.85 & 0.15 & 0.51
        & - & - & - & - & - & - \\
      SD2.1 +\textbf{\frameworkName{}} & \isbest{0.32} & 0.55 & 0.43 & 0.54 & 0.30
        & \isbest{0.51} & 0.99 & 0.69 & 0.20 & 0.71 & 0.15 & \isbest{0.54}
        & \isbest{86.54} & 79.94 & 78.89 & 75.39 & 62.80 & \isbest{69.48} \\
      \midrule
      FLUX.1     & 0.18 & 0.69 & 0.48 & 0.63 & 0.31
        & 0.26 & 0.92 & 0.77 & 0.71 & 0.66 & 0.27 & 0.60
        & 92.30 & 80.55 & 87.74 & 85.55 & 78.40 & 80.63 \\
      FLUX.1 +\textbf{\frameworkName{}} & \isbest{0.30} & 0.83 & 0.59 & 0.71 & 0.32
        & \isbest{0.60} & 0.99 & 0.87 & 0.71 & 0.80 & 0.76 & \isbest{0.76}
        & \isbest{94.12} & 82.98 & 90.53 & 88.30 & 82.80 & \isbest{84.42} \\
      \bottomrule
  \end{tabular}
  }
\end{table*}

\end{document}